\documentclass{article}

\usepackage{bbold}
\usepackage{adjustbox}
\usepackage{multirow}   
\usepackage{microtype}
\usepackage{graphicx}
\usepackage{subfigure}
\usepackage{booktabs} 

\PassOptionsToPackage{hyphens}{url}\usepackage{hyperref}

\usepackage{arxiv}[preprint]

\usepackage{amssymb}
\usepackage{amsmath,amsfonts,bm}
\usepackage{mathtools}
\usepackage{amsthm}

\newcommand{\mF}{\bm{F}}
\newcommand{\mZ}{\bm{Z}}
\newcommand{\mX}{\bm{X}}
\newcommand{\mH}{\bm{H}}
\newcommand{\mR}{\bm{R}}
\newcommand{\mW}{\bm{W}}
\newcommand{\mS}{\bm{S}}
\newcommand{\mB}{\bm{B}}
\newcommand{\vone}{\bm{1}}

\def\vone{{\bm{1}}}

\def\vb{{\bm{b}}}

\def\vh{{\bm{h}}}

\def\vr{{\bm{r}}}
\def\vs{{\bm{s}}}

\def\vx{{\bm{x}}}

\def\vz{{\bm{z}}}

\newcommand{\R}{\mathbb{R}}

\usepackage[capitalize,noabbrev]{cleveref}

\theoremstyle{plain}

\theoremstyle{definition}

\theoremstyle{remark}

\usepackage[textsize=tiny]{todonotes}

\plaintitlerunning{Evaluating Prediction Uncertainty Estimates from BatchEnsemble}

\begin{document}

\twocolumn[
\plaintitle{Evaluating Prediction Uncertainty Estimates from BatchEnsemble}

\plainsetsymbol{equal}{*}

\begin{plainauthorlist}
\plainauthor{Morten Blørstad}{yyy}
\plainauthor{Herman Jangsett Mostein}{comp}
\plainauthor{Nello Blaser}{yyy}
\plainauthor{Pekka Parviainen}{yyy}

\end{plainauthorlist}

\plainaffiliation{yyy}{Department of Informatics, University of Bergen, Bergen, Norway. }
\plainaffiliation{comp}{IBM, Copenhagen, Denmark}

\vskip 0.3in
]

\printAffiliationsAndNotice{}  

\begin{abstract}
Deep learning models struggle with uncertainty estimation. Many approaches are either computationally infeasible or underestimate uncertainty. 
We investigate \textit{BatchEnsemble} as a general and scalable method for uncertainty estimation across both tabular and time series tasks. To extend BatchEnsemble to sequential modeling, we introduce GRUBE, a novel BatchEnsemble GRU cell. We compare the BatchEnsemble to Monte Carlo dropout and deep ensemble models.
Our results show that BatchEnsemble matches the uncertainty estimation performance of deep ensembles, and clearly outperforms Monte Carlo dropout. GRUBE achieves similar or better performance in both prediction and uncertainty estimation. 
These findings show that BatchEnsemble and GRUBE achieve similar performance with fewer parameters and reduced training and inference time compared to traditional ensembles. 
\end{abstract}

\section{Introduction}

Deep learning models are used across various industries and domains, such as weather, finance, energy, and healthcare. Most models provide only point predictions and lack well calibrated uncertainty estimates. For many applications, representing predictive uncertainty is essential and remains a long-standing practical challenge. 

Predictive uncertainty is typically decomposed into \textit{aleatoric uncertainty}, capturing inherent data noise, and \textit{epistemic uncertainty}, reflecting model uncertainty. Epistemic uncertainty can be used to identify out-of-distribution data and distribution shifts to enable reliable decision-making in safety-critical settings. Methods that fail to capture epistemic uncertainty may appear well-calibrated while remaining overconfident on unfamiliar inputs.

Neural networks often fail to quantify predictive uncertainty and tend to produce overconfident predictions \citep{Lakshminarayanan2017}.  Bayesian methods \citep{Graves2011,Welling2011, blundell2015, Izmailov2021, Mirikitani2010, Chien2016, Fortunato2019} overcome overconfidence, but are typically hard to train and computationally expensive. This motivates alternative uncertainty estimation methods that only require minor changes to standard training pipelines. 

Monte Carlo (MC) dropout \cite{Gal2016Dropout} is such a method, but can be overconfident \citep{Lakshminarayanan2017, Izmailov2021}. Deep ensembles have been proposed as another alternative, and have been shown to generate well-calibrated uncertainty estimates \citep{Lakshminarayanan2017}. Later studies found that deep ensembles usually outperform MC dropout on out-of-distribution (OOD) uncertainty benchmarks \citep{Ovadia2019, gustafsson2020}. In practice deep ensembles are limited by their high computational and memory costs. 

BatchEnsemble \citep{Wen2020} mitigates these costs by sharing weights across ensemble members, providing a parameter-efficient alternative to deep ensembles. BatchEnsemble yields strong predictive performance in image classification and machine translation and competitive accuracy and uncertainties as deep ensembles on out-of-distribution data in image classification \citep{Wen2020}. TabM \citep{Gorishniy2025} applied similar ideas to tabular data, focusing on efficiency and predictive performance without considering uncertainty estimates.

\citet{zamyatin2026} have analyzed the calibration and diversity of BatchEnsemble for image classification tasks and found BatchEnsemble to underperform deep ensembles, contradicting results from \citet{Wen2020}.
Except for the work on machine translation \citep{Wen2020}, to the best of our knowledge there is no work on BatchEnsemble for uncertainty quantification on tabular and time series data, or on BatchEnsemble using a recurrent neural network. 
We evaluate the performance of BatchEnsemble in prediction and uncertainty estimation  on tabular regression and classification and on time series forecasting. 

\textbf{Our contribution.}
We empirically investigate if BatchEnsemble provides accurate and reliable uncertainty estimates as a efficient alternative to deep ensembles across different tasks.  
Our main contributions are:
\begin{itemize}
\vspace{-1em}
    \item We demonstrate that BatchEnsemble strikes a balance between MC dropout and deep ensembles, achieving better uncertainty estimates than MC dropout while matching deep ensembles with significantly fewer parameters across multiple tasks and data modalities.
    \item We introduce GRUBE, a BatchEnsemble-based GRU variant, extending BatchEnsemble to time series tasks by integrating it directly into the GRU cells. We show that GRUBE achieves similar or better predictive and uncertainty estimation performance with substantially fewer parameters.
\end{itemize}

BatchEnsemble closely matches or surpasses deep ensembles in uncertainty estimation and predictive performance while being far more parameter-efficient. GRUBE extends these benefits to time-series tasks with minimal cost.

\section{BatchEnsemble For Scalable Predictive Uncertainty Estimation}
\label{sec:batchensemble}

\subsection{Problem setup}
We study supervised learning on tabular and time-series data under a unified notation. For tabular tasks, we observe a dataset \(\mathcal{D}=\{(\vx_i,y_i)\}_{i=1}^N\) with inputs \(\vx_i\in\mathbb{R}^d\) and targets $y_i$. For regression \(y_i\in\mathbb{R}\) and for classification \(y_i\in\{1,\ldots,C\}\). The goal is to learn a predictive distribution \(p(y\mid\vx, \theta)\) with model parameters \(\theta\). 
Time-series observations arrive as an ordered sequence \(\{(\vx_t,y_t)\}_{t=1}^T\). Forecasting conditions on a context window of length \(L\) and predicts the next \(H\) steps through \(p\!\big(y_{t+1:t+H}\mid \vx_{t-L+1:t}, \theta \big)\).

Similar to MC dropout \citep{Gal2016Dropout} and deep ensemble \citep{Lakshminarayanan2017}, we treat BatchEnsemble as a practical Bayesian approximation with minimal modifications to standard training. The predictive distribution
$
p(y \mid \vx) = \int p(y \mid \vx, \theta)\, p(\theta \mid \mathcal{D})\, d\theta$
is approximated by an ensemble size of \(K\) members
\[
 p(y \mid x) \approx \frac{1}{K} \sum_{k=1}^{K} p(y \mid \vx, \theta_k), 
\]
where each \(\theta_k\) corresponds to the parameters of the \(k\)-th BatchEnsemble member, obtained by combining shared weights with member-specific adapters (see Section~\ref{subsec:BE}).

\subsection{BatchEnsemble}
\label{subsec:BE}
Standard deep ensembles are accurate but costly in memory and computation because each member has its own full set of weights. \emph{BatchEnsemble} \citep{Wen2020} shares one weight matrix $\mW\!\in\!\R^{p\times q}$ per layer $\ell$, where $p$ and $q$ are the input and output size. Each member $k$ is given only three small vectors: input adapter $\vr_k\!\in\!\R^{p}$, output adapter $\vs_k\!\in\!\R^{q}$, and optional bias $\vb_k\!\in\!\R^{q}$. These adapters modulate the shared weights through element-wise scaling of the input and output activations, adding only $p{+}q$ parameters per member (plus $q$ if a bias is used), instead of $pq$ like full ensemble members would.

For a single input $\vz_{\ell}^{(k)}\in\!\R^{p}$ and nonlinearity $\phi$, member $k$ computes
\[
\vz_{\ell+1}^{(k)}=\phi\!\Big(\big(\mW^\top(\vz_{\ell}^{(k)}\odot\vr_k)\big)\odot\vs_k+\vb_k\Big).
\]

To vectorize computations, we repeat inputs for the first layer $K$ times to get $\mZ_{0}\!\in\!\R^{K\times p}$. Per-member vectors $\vr_k$, $\vs_k$ and $\vb_k$ are stacked into matrices $\mR\!\in\!\R^{K \times p}$, $\mS\!\in\!\R^{K \times q}$ and $\mB\!\in\!\R^{K \times q}$. The vectorized forward pass becomes
\[
\mZ_{\ell +1}=\phi\!\Big(\big((\mZ_{\ell}\odot\mR)\,\mW\big)\odot\mS+\mB\Big) \in\R^{K\times  q}.
\]

\subsection{Feed-forward Neural Network with BatchEnsemble}

For feed-forward networks, each linear layer 
is replaced by its BatchEnsemble counterpart as described above. 
This substitution allows multiple ensemble members to be trained and evaluated in parallel using shared weights, while maintaining the same overall architecture and training pipeline.

\textbf{Training.}
We optimize the ensemble by minimizing the average per-member negative log-likelihood. Training is fully vectorized; a single forward/backward pass computes all members’ losses simultaneously. Shared parameters receive aggregated gradients across members, whereas member-specific parameters are updated from their member’s loss.

For classification, we use the categorical distribution
$
\mathcal{L}_{\text{NLL}} = -\sum_{c=1}^{C} \mathbb{1}[y=c] \log p(y=c \mid \vx, \theta)
$
and for regression, we use the Gaussian distribution $
\mathcal{L}_{\text{NLL}} = 
\frac{1}{2}\log\sigma_{\theta}^2(x) 
+ \frac{(y-\mu_\theta(x))^2}{2\sigma_{\theta}^2(x)}.$ 
For regression tasks we include a heteroscedastic head, so the model outputs $(\mu_\theta(x), \log\sigma_\theta^2(x))$.

\textbf{Inference.}
For classification, the predictive distribution corresponds to averaging the predicted class probabilities over the ensemble members. For regression, the predictive distribution results in a mixture of Gaussian distributions. Following \cite{Lakshminarayanan2017}, we approximate this mixture with a single Gaussian whose mean and variance are given by
\[
\begin{aligned}
\mu^*(x) &= \frac{1}{K}\sum_{k=1}^{K}\mu_{\theta_k}(x),\\
\sigma^{2*}(x) &= \frac{1}{K}\sum_{k=1}^{K}\big[\sigma_{\theta_k}^2(x) + \mu_{\theta_k}^2(x)\big] - \mu^{*}(x)^2.
\end{aligned}
\]

\section{Recurrent Neural Network with BatchEnsemble}
Recurrent architectures can be extended with BatchEnsemble, enabling parameter-efficient ensemble for sequential data such as time series. 
A standard recurrent neural network (RNN) updates its hidden state as $\vh_t = \tanh\!\left(\mW[\vx_t, \vh_{t-1}] + \vb\right)$, where $\vx_t$ is the input and $\vh_{t-1}$ the previous hidden state. 
In the BatchEnsemble variant, the output of recurrent layer is the hidden states for all $K$ members $\mH_t \in \R^{K\times  q}$. To construct the ensemble input at time $t$, the current input $\vx_t$ is repeated $K$ times to form $\mX_t \in \R^{K\times p}$ and concatenated with the previous ensemble hidden state $\mH_{t-1}$. The shared weights are modulated by member-specific adapters, yielding
\[
\mH_t = \tanh\!\left(\left(\left([\mX_t, \mH_{t-1}] \odot \mR\right)\mW\right) \odot \mS + \mB\right),
\]
where $\mW\in\mathbb{R}^{(p+q)\times q}$ is the shared weights and  $\mR\!\in\!\R^{K\times(p+q)}$, $\mS,\mB\!\in\!\R^{K\times q}$ contain stacked per-member adapters that are broadcast across the batch dimension. 
This allows all ensemble members to be computed in parallel within a single forward pass, keeping efficiency similar to a single model while capturing diverse function realizations. 

\subsection{GRU with BatchEnsemble (GRUBE)}

\begin{figure}[h]
    \centering
    \includegraphics[width=0.99\linewidth]{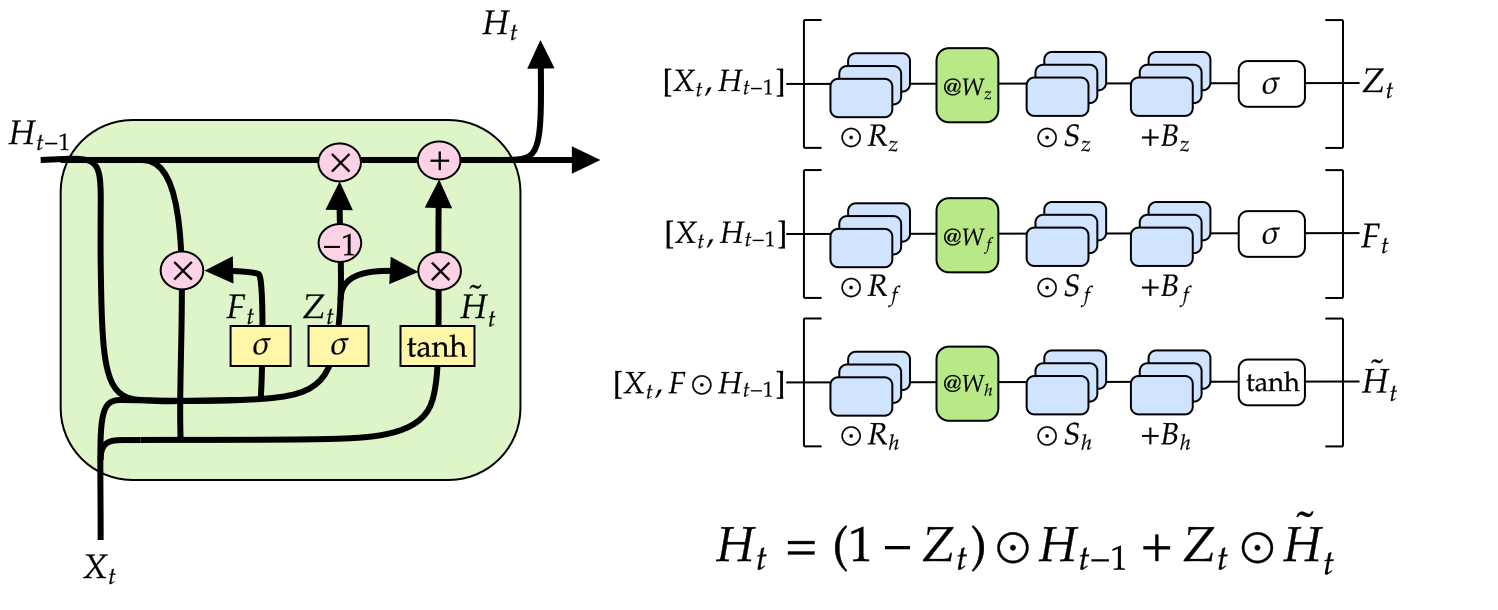}
    \vspace{-1em}\\
    \caption{GRU with BatchEnsemble (GRUBE). The architecture combines the standard GRU architecture with the BatchEnsemble formalism applied to the weights \(W_z, W_f\), and \(W_h\).}
    \label{fig:grube}
\end{figure}

We introduce GRUBE (Figure \ref{fig:grube}), a GRU \cite{Cho2014} variant that employs BatchEnsemble to quantify uncertainty in time-series forecasting tasks. GRUBE accounts for both aleatoric and epistemic uncertainty by replacing all linear transformations from GRU with shared-weight ensembles, 
\begin{align*}
\mF_t &= \sigma\!\left(\big(\big([\mX_t,\mH_{t-1}] \odot \mR_f\big)\mW_f\big) \odot \mS_f + \mB_f\right),\\
\mZ_t &= \sigma\!\left(\big(\big([\mX_t,\mH_{t-1}] \odot \mR_z\big)\mW_z\big) \odot \mS_z + \mB_z\right),\\
\tilde{\mH}_t &= \tanh\!\left(\big(\big([\mX_t,\mF_t \odot \mH_{t-1}] \odot \mR_h\big)\mW_h\big) \odot \mS_h + \mB_h\right),\\
\mH_t &= \big(\vone - \mZ_t\big) \odot \mH_{t-1} + \mZ_t \odot \tilde{\mH}_t .
\end{align*}
This formulation creates an efficient ensemble of recurrent models that share most parameters, significantly reducing memory and compute costs while maintaining predictive diversity for uncertainty estimation. 

\textbf{Inference.}
For time-series forecasting, uncertainty compounds as predictions are rolled forward. Each prediction depends on previous uncertain outputs, causing both aleatoric and epistemic uncertainty to accumulate. To account for this, we use ancestral sampling. For each ensemble member \(k\) and sample path \(s\), we draw predictions recursively as
\[
\hat{y}^{(t+h)}_{k,s} =
\hat{\mu}^{(t+h)}_{k}\!\big(\hat{y}^{(t:h-1)}_{k,s}\big)
+ \hat{\sigma}^{(t+h)}_{k}\!\big(\hat{y}^{(t:h-1)}_{k,s}\big)\,\epsilon^{(t+h)}_{k,s}, \\
\]
where \(\epsilon^{(t+h)}_{k,s} \sim \mathcal{N}(0,1)\).

We then aggregate over all $K$ members and $S$ sample paths per member to get
\[
\begin{aligned}
\hat{\mu}^{(t+h)} &= \frac{1}{KS}\sum_{k,s} \hat{y}^{(t+h)}_{k,s}, \\
\hat{\sigma}^2\!\big(Y^{(t+h)}\big) &=
\frac{1}{KS}\sum_{k,s}\big(\hat{y}^{(t+h)}_{k,s}-\hat{\mu}^{(t+h)}\big)^2.
\end{aligned}
\]

\section{Experiments}

The goal of our experiments is to compare BatchEnsemble, deep ensemble, and MC dropout in terms of predictive performance and uncertainty estimation performance, in-distribution and under distribution shift. We assess uncertainty estimates from complementary perspectives, including quality of the full predictive distribution, calibration, sensitivity to distribution shift, and if uncertainty estimates are informative for ranking prediction difficulty. 
We further analyze uncertainty behavior through aleatoric–epistemic decomposition. We consider uncertainty estimates reliable if models perform well across all evaluated criteria.

\subsection{Experimental Setup}

\textbf{Datasets.} 
We evaluate BatchEnsemble on various datasets to get an comprehensive assessment of the uncertainty calibration across modalities and tasks (\cref{tab:data_overview}). 

To evaluate robustness under distribution shift, we construct shifted variants of the tabular datasets by excluding samples in the lower or upper q-quantile of two selected numerical features from training and assigning them to the test set. The numerical features are selected based on feature importance using a generalized linear model and we use $q=0.025$. See Appendix \ref{appendix:tails} for more details. The time-series datasets already exhibits distributional shifts due to its temporal structure and is evaluated without additional modifications.

\textbf{Setup.}
For all tasks, we split data into train and test using a 80/20 split. Table \ref{tab:params_size} shows the number of parameters for each model and dataset. Results over five runs with different seeds are reported as mean~$\pm$~standard error.

For tabular data, we shuffle the data before splitting, whereas for times series we split chronologically. Data is scaled to $[0,1]$. We train the models by minimizing the \emph{negative log likelihood} (NLL) with a fixed set of hyperparameters (see Appendix~\ref{appendix:setup}). 
For time-series tasks, we use a context window of $L=12$ and a prediction horizon of $H=5$. Training is fully autoregressive (\emph{no teacher forcing}), and the objective is a \emph{multi-step NLL}, averaged over all forecast steps. 
At test time, multi-step forecasts are generated through ancestral sampling with 2000 samples, where each ensemble member produces trajectories for the next $H$ steps.

\textbf{Uncertainty estimation metrics.}
Uncertainty estimation performance is divided into proper scoring rules, which assess the full predictive distribution, and calibration metrics, which measure agreement between predicted probabilities or intervals and empirical frequencies. We use NLL and the Brier score as proper scoring rules, where
$$\mathrm{Brier} = \frac{1}{N}\sum_{i=1}^{N} \sum_{c=1}^{C} 
\big( p(y_i = c \mid \vx_i, \theta) - \mathbb{1}[y_i = c] \big)^2.$$
For regression, calibration is evaluated using central prediction intervals at nominal coverages $\mathcal{G}=\{\gamma_j\}_{j=1}^{J}$, evenly spaced in $[0.025,0.975]$, where $\gamma_j$ denotes the target coverage probability of the predictive interval. Let $\hat{\gamma}_j$ denote the empirical coverage at level $\gamma_j$. We compute $\text{RMSCE}=\sqrt{\frac{1}{J}\sum_{j=1}^{J}(\hat{\gamma}_j-\gamma_j)^2}$, and miscalibration area $\int_{0}^{1}\lvert\hat{\gamma}(\gamma)-\gamma\rvert\,d\gamma$, approximated numerically using the trapezoidal rule over $\mathcal{G}$.

For classification, calibration is evaluated using the expected calibration error (ECE),
\[
\mathrm{ECE}=\sum_{m=1}^{M}\frac{|B_m|}{N}\,\big|\mathrm{acc}(B_m)-\mathrm{conf}(B_m)\big|.
\]
Predictions are partitioned into $M=15$ uniform confidence bins on $[0,1]$, where $B_m$ denotes the set of samples whose predicted confidence $\hat{p}_i=\max_c p(y=c\mid \vx_i,\theta)$ falls into bin $m$. The quantities $\mathrm{acc}(B_m)$ and $\mathrm{conf}(B_m)$ denote empirical accuracy and average predicted confidence of samples in bin $m$, respectively.

\textbf{Uncertainty decomposition.}
We decompose predictive uncertainty into aleatoric and epistemic components \citep{Depeweg2018}. 
For regression, the predictive variance $\sigma^{2*}(\vx)$ is decomposed as
\[
\sigma^{2*}(\vx)
=
\underbrace{\frac{1}{K}\sum_{k=1}^{K}\sigma^2_{\theta_k}(\vx)}_{\text{aleatoric}}
+
\underbrace{\frac{1}{K}\sum_{k=1}^{K}\big(\mu_{\theta_k}(\vx)-\mu^*(\vx)\big)^2}_{\text{epistemic}}.
\]

For classification, predictive uncertainty is decomposed using the ensemble class probabilities 
$\bar{p}(y=c\mid \vx)=\frac{1}{K}\sum_{k=1}^{K} p(y=c\mid \vx, \theta_k)$. 
Total uncertainty is measured by the predictive entropy
\[
H(\bar{p}) = -\sum_{c=1}^{C}\bar{p}(y=c\mid \vx)\log \bar{p}(y=c\mid \vx).
\]
We define the average member entropy as
\[
\overline{H} = \frac{1}{K}\sum_{k=1}^{K} H\!\big(p(y\mid \vx,\theta_k)\big),
\]
where $H(p) = -\sum_{c=1}^{C} p(y=c)\log p(y=c)$.
Aleatoric uncertainty is given by $\overline{H}(p)$, and epistemic uncertainty by $$H(\bar{p}) - \overline{H}.$$
\textbf{Selective Prediction Evaluation.}
In addition to prediction and uncertainty estimation performance, we evaluate whether the uncertainty estimates are informative for ranking prediction difficulty using selective prediction \citep{Herbei2006, Geifman2017,Lakshminarayanan2017}. 

The model outputs predictions and the corresponding uncertainty scores, $u(\vx)$ for all samples. For classification, uncertainty is measured using the predictive entropy of the class probability vector. For regression, uncertainty is given by standard deviation of the predictive distribution. For time-series forecasting, uncertainty is computed as the average predictive standard deviation over the forecasting horizon. 

For a nominal coverage level $\gamma \in (0,1]$, we select a subset 
$\mathcal{S}_\gamma=\{i:u(\vx)\leq \tau_\gamma\},$
where $\tau_\gamma$ is chosen such that $ \lvert\mathcal{S}_\gamma\rvert = \lceil\gamma N\rceil$. Performance is evaluated on  $\mathcal{S}_\gamma$. This results in a coverage–performance curve that shows how predictive performance changes as increasingly uncertain samples are included.

\subsection{Experimental Results}

\subsubsection{Tabular Regression}

\begin{figure}[h]
    \centering
    \includegraphics[width=0.99\linewidth]{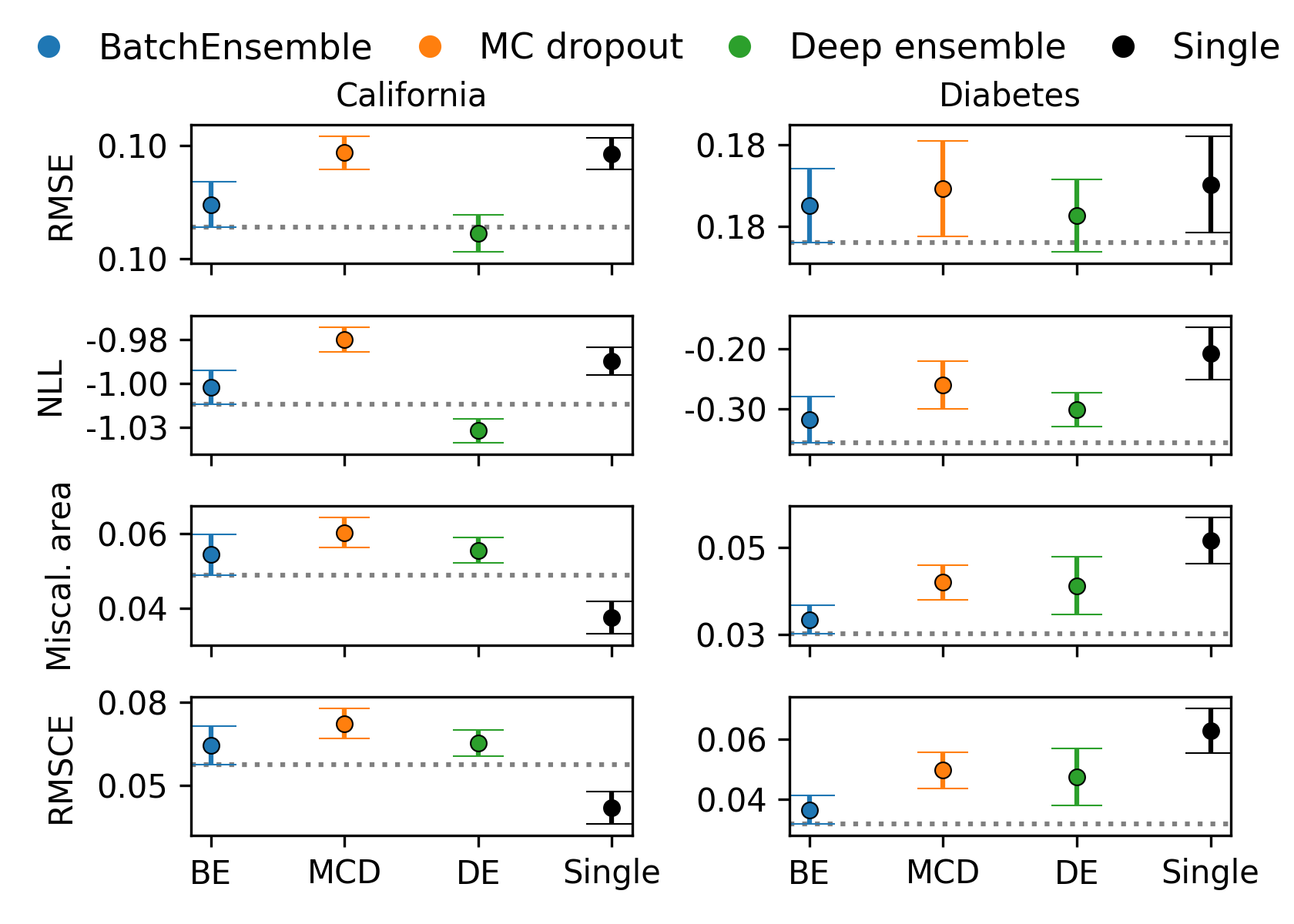}
    \vspace{-1em}\\
    \caption{Regression results across tabular datasets. For each metric it shows the mean $\pm$ standard error. The dotted line shows the lower limit of BatchEnsemble.}
    \label{fig:reg_main}
\end{figure}

\textbf{In-distribution.}
Figure \ref{fig:reg_main} shows that on tabular regression tasks, BatchEnsemble is tied for best RMSE on both datasets under the $\mathrm{mean} \pm \mathrm{SE}$ overlap rule, matching deep ensembles on California and all methods on Diabetes. For uncertainty estimation, deep ensemble achieves the best NLL on California, closely followed by BatchEnsemble. On Diabetes there is a tie between BatchEnsemble, deep ensemble and MC dropout. Single model has the best calibration scores on California, whereas BatchEnsemble and deep ensemble achieve the best calibration scores on Diabetes. A closer inspection reveals that all models consistently overestimates uncertainty on California across all coverage levels (Figure \ref{fig:nom_vs_emp_cov_reg}, Appendix \ref{appendix:nom_vs_emp}). Table \ref{tab:unc_id_shift_delta_reg} shows all models have similar aleatoric uncertainty, but BatchEnsemble, deep ensemble and MC dropout additionally capture epistemic uncertainty, which leads to higher calibration errors.

\begin{figure}[h]
    \centering
    \includegraphics[width=0.99\linewidth]{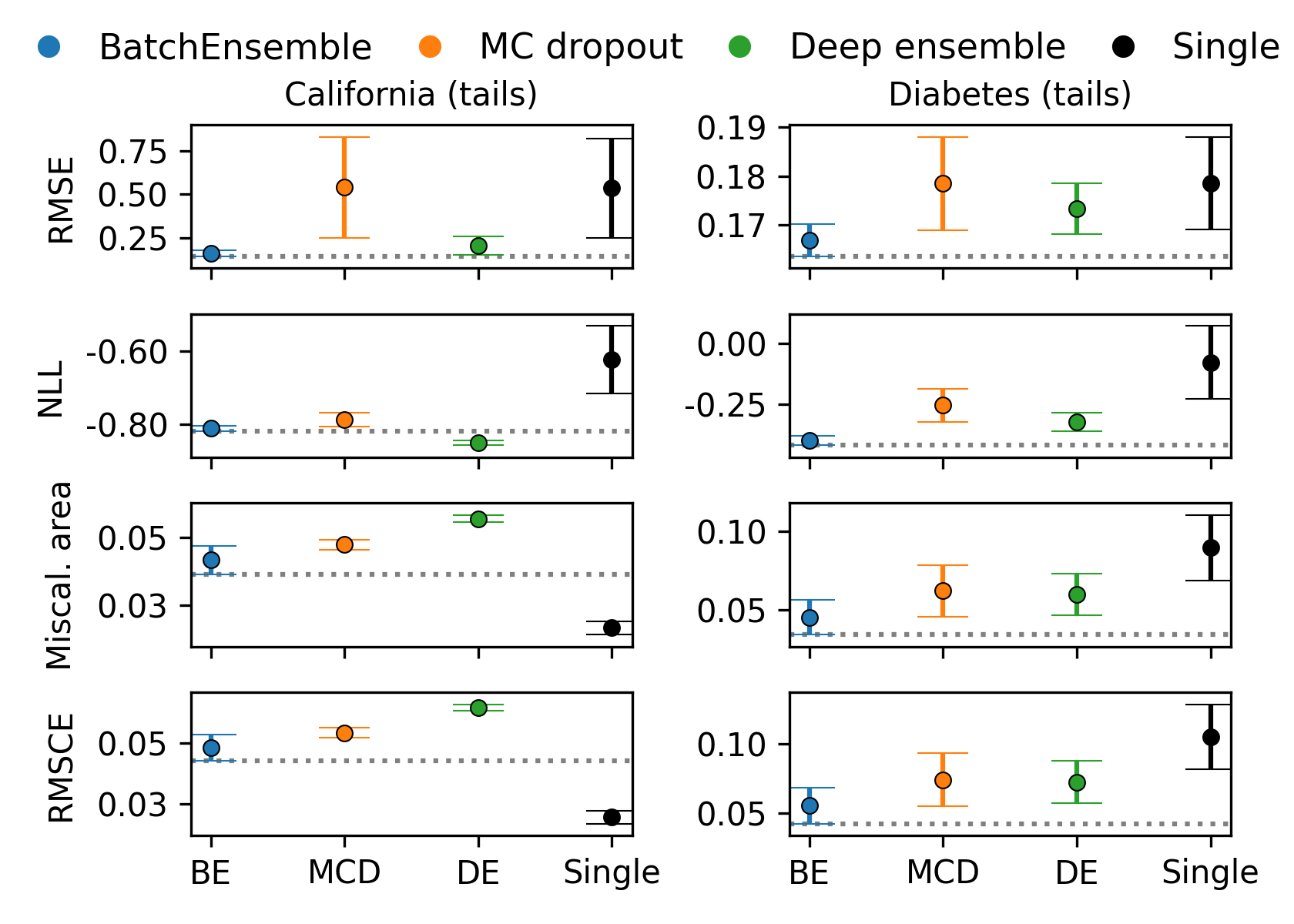}
    \vspace{-1em}\\
    \caption{Regression results across tabular datasets with distribution shift. For each metric it shows the mean $\pm$ standard error. The dotted line shows the lower limit of BatchEnsemble.}
    \label{fig:reg_tails}
\end{figure}

\textbf{Distribution shift.}
Figure~\ref{fig:reg_tails} shows how the models performs under distribution shift. On California, BatchEnsemble and deep ensemble remain robust in RMSE ($\approx0.097 \to 0.18$), while MC dropout and the single model degrade substantially ($\approx0.1 \to 0.54$). For Diabetes the predictive performance is still a tie and similar RMSE values. For uncertainty estimation, deep ensemble still achieves the best NLL, closely followed by BatchEnsemble, and single model still achieves the best calibration metrics on the California dataset. On the Diabetes dataset BatchEnsemble achieves the best NLL, closely followed by deep ensemble and MC dropout. There is a tie between BatchEnsemble, deep ensemble and MC dropout on the calibration metrics.

\textbf{Uncertainty decomposition.}
Table~\ref{tab:unc_id_shift_delta_reg} shows the uncertainty decomposition on the in-distribution and distribution shift and the change. On in-distribution data, all models exhibit similar levels of aleatoric uncertainty on both datasets. For California, BatchEnsemble and deep ensemble have nearly identical and very small epistemic uncertainty, while MC dropout exhibits substantially higher epistemic uncertainty.

\begin{table}[tbh]
\centering
\textbf{\scriptsize Dataset: \texttt{California}}\\
\begin{adjustbox}{max width=0.48\textwidth}
\begin{tabular}{lccc|ccc|ccc}
\toprule
 & \multicolumn{3}{c|}{ID} & \multicolumn{3}{c|}{Shift} & \multicolumn{3}{c}{$\Delta$ (Shift $-$ ID)} \\
\cmidrule(lr){2-4} \cmidrule(lr){5-7} \cmidrule(lr){8-10}
Model & total & aleatoric & epistemic & total & aleatoric & epistemic & total & aleatoric & epistemic \\
\midrule
BatchEnsemble & 0.013 & 0.012 & 0.001 & 0.111 & 0.038 & 0.073 & +0.098 & +0.026 & +0.072 \\
MC dropout & 0.024 & 0.011 & 0.013 & 0.226 & 0.050 & 0.176 & +0.202 & +0.039 & +0.163 \\
Deep ensemble & 0.011 & 0.010 & 0.001 & 0.417 & 0.031 & 0.386 & +0.406 & +0.021 & +0.385 \\
Single & 0.011 & 0.011 & 0.000 & 0.044 & 0.044 & 0.000 & +0.033 & +0.033 & +0.000 \\
\bottomrule
\end{tabular}
\end{adjustbox}
\vspace{0.01em}

\textbf{\scriptsize Dataset: \texttt{Diabetes}}

\begin{adjustbox}{max width=0.48\textwidth}
\begin{tabular}{lccc|ccc|ccc}
\toprule
 & \multicolumn{3}{c|}{ID} & \multicolumn{3}{c|}{Shift} & \multicolumn{3}{c}{$\Delta$ (Shift $-$ ID)} \\
\cmidrule(lr){2-4} \cmidrule(lr){5-7} \cmidrule(lr){8-10}
Model & total & aleatoric & epistemic & total & aleatoric & epistemic & total & aleatoric & epistemic \\
\midrule
BatchEnsemble & 0.029 & 0.028 & 0.001 & 0.026 & 0.025 & 0.001 & -0.003 & -0.003 & +0.000 \\
MC dropout & 0.053 & 0.026 & 0.027 & 0.055 & 0.027 & 0.029 & +0.002 & +0.001 & +0.002 \\
Deep ensemble & 0.026 & 0.026 & 0.000 & 0.027 & 0.026 & 0.001 & +0.001 & +0.000 & +0.001 \\
Single & 0.025 & 0.025 & 0.000 & 0.025 & 0.025 & 0.000 & +0.000 & +0.000 & +0.000 \\
\bottomrule
\end{tabular}
\end{adjustbox}
\vspace{-1em}
\caption{Uncertainty decomposition for tabular regression under in-distribution (ID) and shifted data. $\Delta$ denotes Shift $-$ ID.}
\label{tab:unc_id_shift_delta_reg}
\end{table}

Under distribution shift, the model behavior differs. On California, epistemic uncertainty increases sharply for the ensemble-based methods and MC dropout, accounting for most of increase in total uncertainty. Deep ensemble has the largest epistemic increase, followed be MC dropout and BatchEnsemble. On Diabetes, there is almost no change in the uncertainty decomposition, with negligible changes in epistemic and total uncertainty.

\begin{figure}[h]
    \centering
    \includegraphics[width=0.9\linewidth]{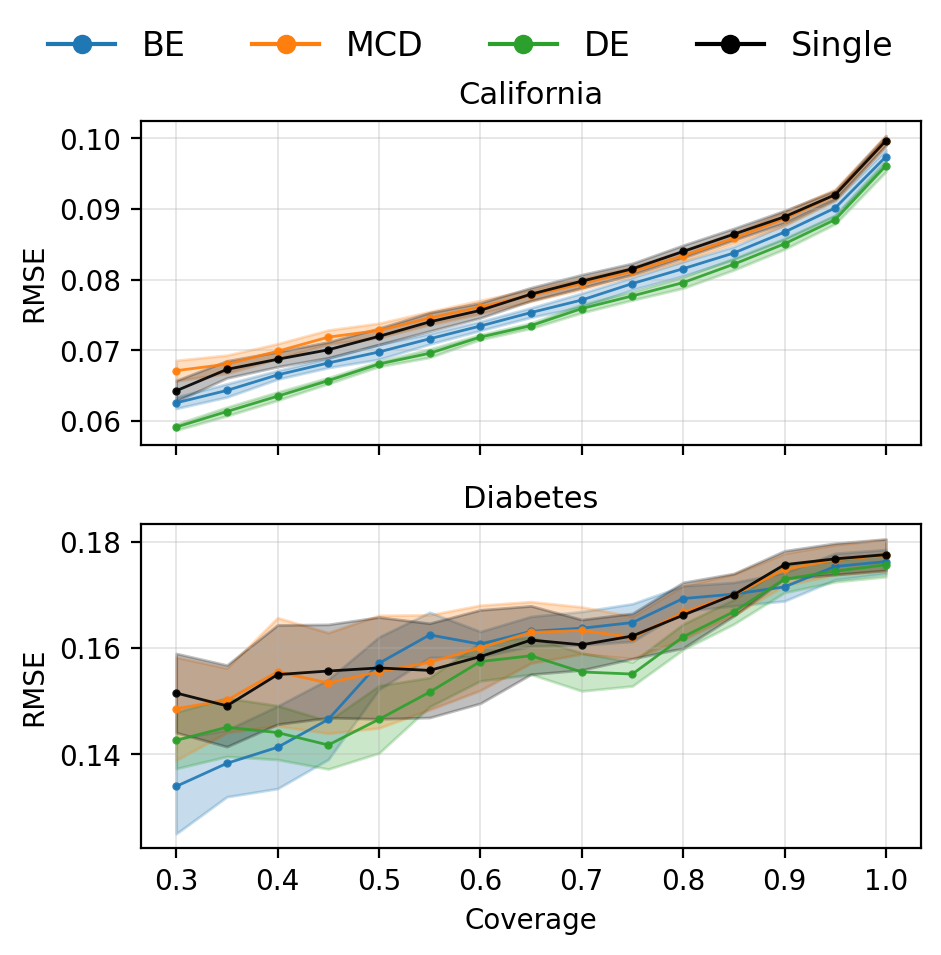}
    \vspace{-1em}\\
    \caption{Selective prediction evaluation on tabular regression datasets.}
    \label{fig:selective_reg}
\end{figure}

\textbf{Selective prediction evaluation.}
Figure \ref{fig:selective_reg} shows predictive performance as a function of uncertainty, where coverage denotes the fraction of most certain samples. BatchEnsemble is competitive with deep ensemble across datasets, with RMSE increasing monotonically with coverage, indicating that more uncertain samples are harder to predict. Both methods consistently provide better rankings than MC dropout and the single model.
On California, deep ensemble performs better than BatchEnsemble at low coverage (top 30–65\% most certain samples). On Diabetes, BatchEnsemble performs best at very low coverage (top 30–35\%), while deep ensemble performs better at intermediate coverage (top 45–80\%). Similar trends holds under distributions shift (Figure~\ref{fig:selective_reg_tails_all}, Appendix~\ref{appendix:selective}).

\subsubsection{Tabular Classification}

\begin{figure}[h]
    \centering
    \includegraphics[width=0.99\linewidth]{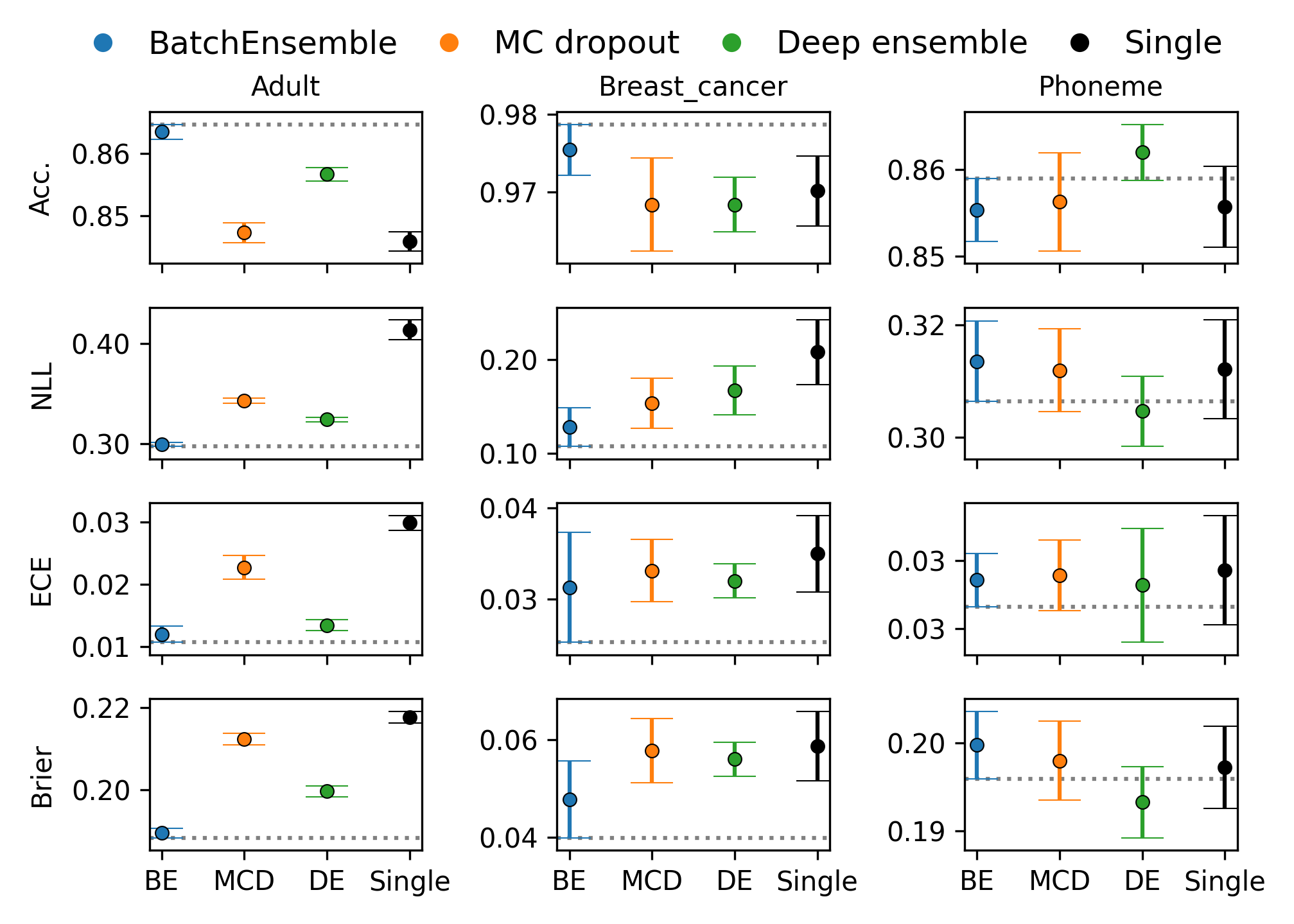}
    \vspace{-1em}\\
    \caption{Classification results across tabular datasets. The dotted line shows the lower limit (upper for accuracy) of BatchEnsemble.}
    \label{fig:clf_main}
\end{figure}

\textbf{In-distribution.}
In-distribution, BatchEnsemble achieves the highest accuracy on Adult. On Breast Cancer, accuracy is tied between BatchEnsemble, MC dropout, and the single model, with deep ensemble slightly worse. On Phoneme, all four methods achieve similar accuracy (\cref{fig:clf_main}). In terms of NLL, BatchEnsemble achieves the best performance on Adult, while BatchEnsemble, MC dropout, and deep ensemble are tied best on Breast Cancer. On Phoneme, all methods are tied. For the Brier score, BatchEnsemble achieves the best performance on Adult, while all methods tied on Breast Cancer and Phoneme. ECE on Adult is tied between BatchEnsemble and deep ensemble, and all methods achieve similar ECE on Breast Cancer and Phoneme.

\textbf{Distribution shift.}
Under the tail-based distribution shift, predictive performance remains unchanged with BatchEnsemble being the best on Adult and a tie between all models on Breast cancer and Phoneme. 

\begin{figure}[h]
    \centering
    \includegraphics[width=0.99\linewidth]{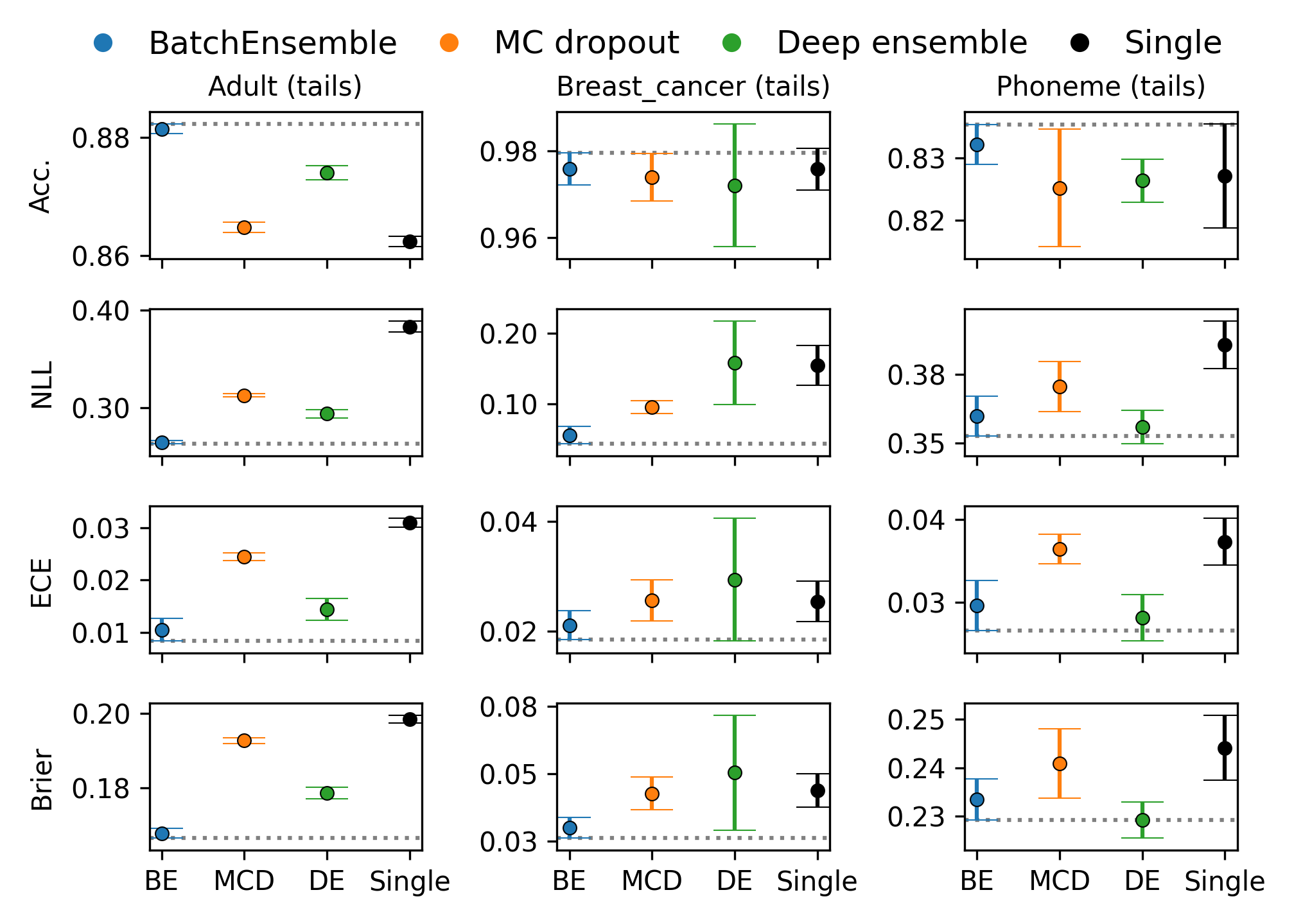}
    \vspace{-1em}\\
    \caption{Classification results across tabular datasets with distribution shift. The dotted line shows the lower limit (upper for accuracy) of BatchEnsemble.}
    \label{fig:clf_tails}
\end{figure}

Differences between methods in uncertainty estimation increase compared to the in-distribution setting. On Adult, BatchEnsemble is best across all uncertainty metrics, followed by deep ensemble and MC dropout. On Breast Cancer, BatchEnsemble gets better NLL than the other methods, while all models remain tied in ECE. Brier score changes from a four-way tie in-distribution to a tie between BatchEnsemble and deep ensembles under distribution shift. On Phoneme, differences between the ensemble-based methods and MC dropout increase, with BatchEnsemble and deep ensemble best on both ECE and Brier (\cref{fig:clf_tails}).   

\textbf{Uncertainty decomposition.}
Appendix Table \ref{tab:unc_id_shift_delta_clf} shows uncertainty decomposition for tabular classification under in-distribution and shifted data. On Adult and Breast Cancer, shifts induce no or small changes in epistemic uncertainty, while Phoneme exhibits substantially larger increases across all models. On Adult under distribution shift, deep ensemble exhibits the highest epistemic uncertainty in absolute values, followed by MC dropout, with BatchEnsemble substantially lower. In contrast, on Breast Cancer, BatchEnsemble exhibits highest epistemic uncertainty, followed by deep ensemble and MC dropout. On shifted Phoneme, BatchEnsemble shows the largest increase in epistemic uncertainty, while MC dropout attains the highest epistemic uncertainty, followed by BatchEnsemble, and deep ensemble.

\begin{figure}[h]
    \centering
    \includegraphics[width=0.9\linewidth]{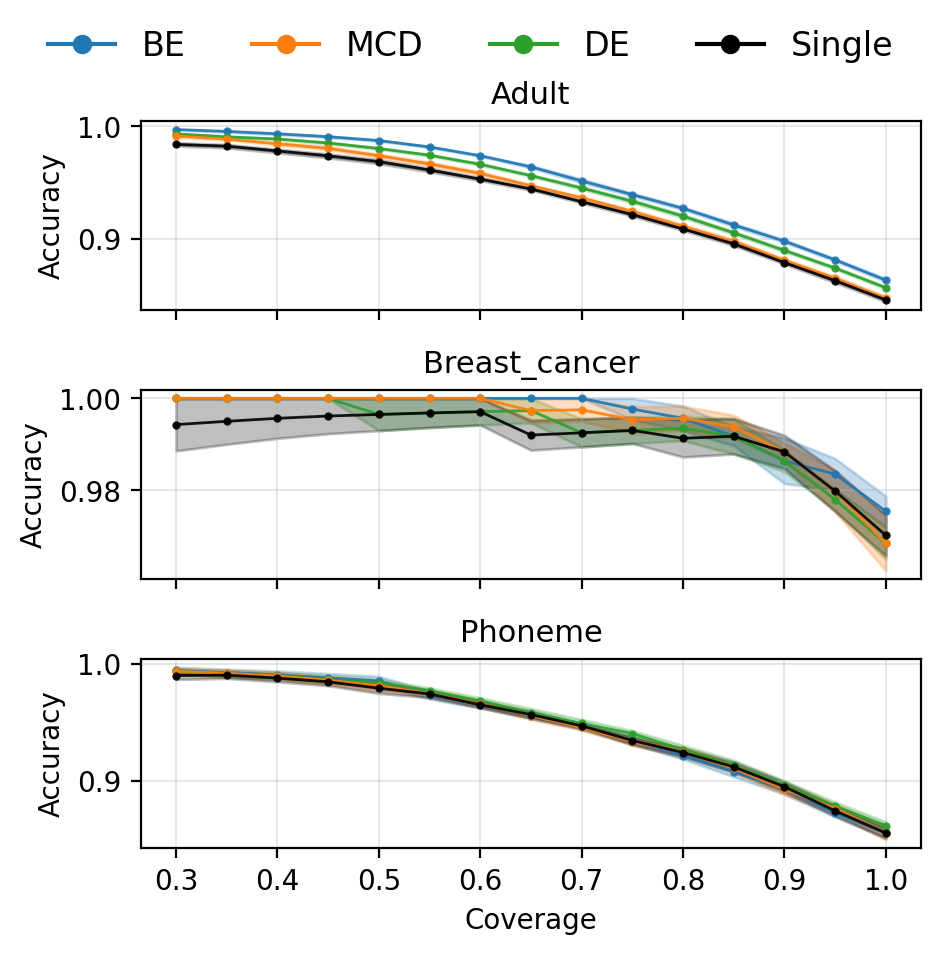}
    \vspace{-1em}\\
    \caption{Selective prediction evaluation on classification datasets.}
    \label{fig:selective_clf}
\end{figure}

\textbf{Selective prediction evaluation.}
Figure~\ref{fig:selective_clf} shows the selective prediction evaluation for in-distribution classification datasets. On the Adult dataset, we find that BatchEnsemble has the strongest predictive accuracy across all coverage, followed by deep ensemble. MC dropout is better than the Single model on the lower levels of coverage, but they are more or less tied as the coverage grows. On the Breast Cancer and Phoneme datasets, all models performs similar. 

Under distribution shift (Figure~\ref{fig:selective_clf_tails_all}, Appendix~\ref{appendix:selective}), similar trends hold on Adult and Phoneme, with clearer separation at lower coverages between models that capture epistemic uncertainty and the single model. On Diabetes dataset, BatchEnsemble are better than deep ensemble on coverage levels 70-85\%. For the other coverages, BatchEnsemble and deep ensemble performs similar. 

\subsubsection{Time-Series Regression}

\begin{figure}[h]
    \centering
    \includegraphics[width=0.99\linewidth]{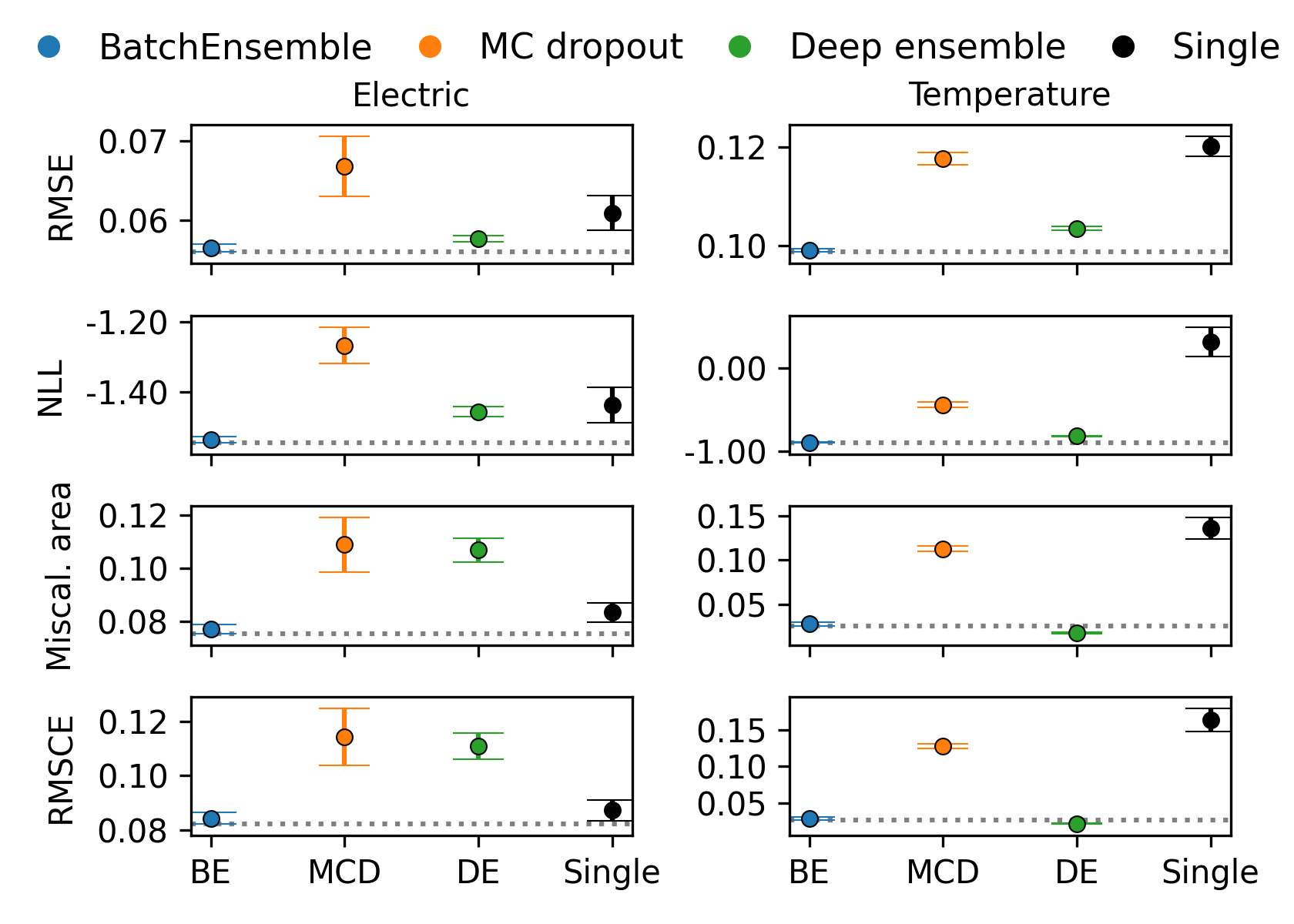}
    \vspace{-1em}\\
    \caption{Regression results across times-series datasets. For each metric it shows the mean $\pm$ standard error. The dotted line shows the lower limit of BatchEnsemble.}
    \label{fig:ts_main}
\end{figure}

On time-series forecasting tasks, BatchEnsemble achieves the best predictive performance on both datasets, closely followed by deep ensemble. MC dropout and Single model are tied for worst predictive performance on both datasets. Detailed results are shown in \cref{fig:ts_main}.

In terms of uncertainty estimation, BatchEnsemble achieves the lowest NLL on both Electric and Temperature. MC dropout has worst NLL on Electric, but is better than Single model on Temperature. Calibration results show that BatchEnsemble achieves the best miscalibration area on Electric and is tied with the single model in RMSCE, while deep ensemble achieves the best calibration on Temperature, with BatchEnsemble as a close second. MC dropout performs similar to deep ensemble in calibration on Electric and similar to Single model on Temperature.

\begin{figure}[h]
    \centering
    \includegraphics[width=0.9\linewidth]{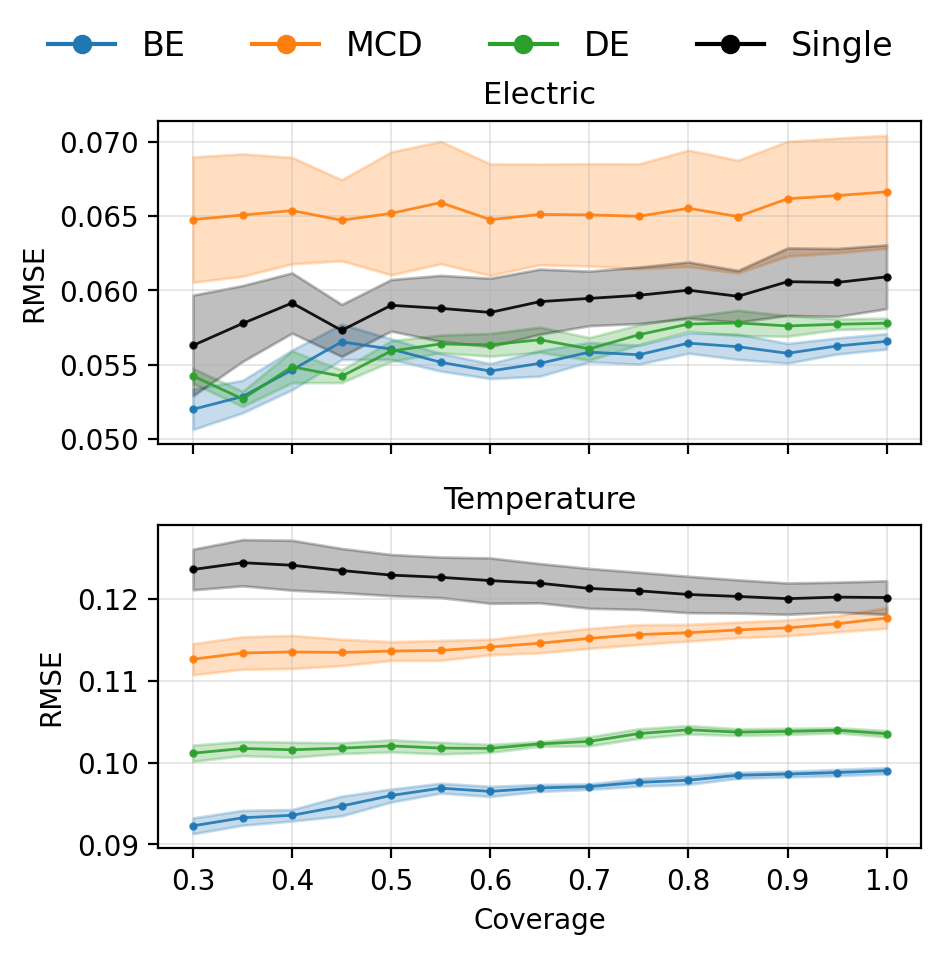}
    \vspace{-1em}\\
    \caption{Selective prediction evaluation on the time-series datasets.}
    \label{fig:selective_reg_temperature}
\end{figure}

\textbf{Selective prediction evaluation.}
Selective evaluation for time-series forecasting is shown in Figure~\ref{fig:selective_reg_temperature}. For BatchEnsemble, MC dropout, and deep ensembles, RMSE generally increases with coverage on both datasets, i.e., performance decreases as less-certain samples are included. This also holds for the single model on Electric. But on Temperature it shows the opposite trend, indicating a mismatch between uncertainty ranking and forecasting difficulty.

On Electric, BatchEnsemble performs best at higher coverage (top 60–100\% most certain samples) and at lower coverage (top 30–60\%) BatchEnsemble is tied best with deep ensemble. MC dropout is consistently worse across coverage levels. 
On Temperature, BatchEnsemble achieves the lowest RMSE across the entire coverage range. Deep ensembles are consistently second-best, while MC dropout and the single model have higher RMSE throughout.

\subsection{Ablation studies}
\label{sec:ablation}

\subsubsection{Orthogonal Initialization}

We explored orthogonal initialization and orthogonal regularization of the BatchEnsemble adapter parameters to encourage decorrelation among ensemble members and improve diversity (Figures \ref{fig:be_init_ablation_reg} and \ref{fig:be_init_ablation_class}, Appendix \ref{appendix:init}). We found that this approach did not provide benefits over the simpler random sign initialization of \citet{Wen2020}.

\subsubsection{BatchEnsemble Adapters}

BatchEnsemble induces ensemble diversity through three per-member adapters: input ($\mR$), output ($\mS$), and bias ($\mB$). 
The relative importance of these components is not well understood. In particular, it is unclear if all three are required to obtain the benefits of BatchEnsemble. We perform an adapter-wise ablation in which we evaluate all possible combinations of ${\mR,\mS,\mB}$. 
All variants share the same base architecture, initialization, and training procedure. 

\begin{figure}[h]
    \centering
    \includegraphics[width=0.99\linewidth]{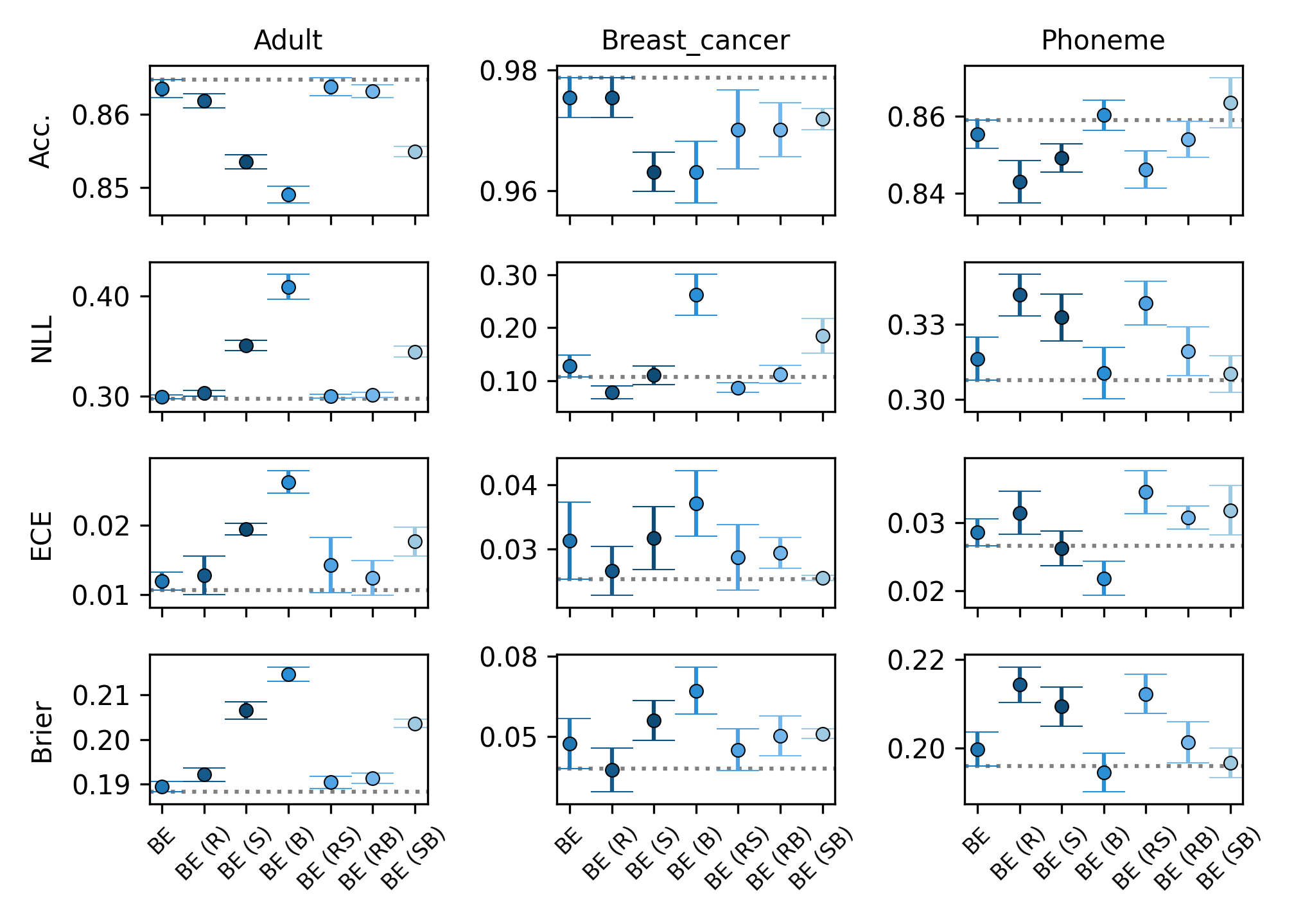}
    \vspace{-1em}\\
    \caption{Adapter-wise ablation on classification datasets.}
    \label{fig:adapter_ablation_clf}
\end{figure}

Figure~\ref{fig:adapter_ablation_clf} below and Figure \ref{fig:adapter_ablation_reg} in Appendix \ref{appendix:adapters} show that while some variants exhibit stronger performance than others, none consistently match or outperform the full BatchEnsemble (BE) model with all adapters included. This means that all adapters are needed. 

\subsubsection{GRU Gates}
We study whether BatchEnsemble must be applied to all GRU gates via a gate-wise ablation, selectively enabling it on the update (Z), reset (F), and candidate (C) gates, testing all possible configurations, including the full model ($CZF$). All variants share the same architecture and training setup. This isolates the contribution of each gate to performance and uncertainty estimation. We found that none of the reduced variants consistently matches or outperforms the full version (\cref{fig:be_gru_ablation}, Appendix \ref{appendix:gru}). Applying BatchEnsemble to all gates provides the most reliable results. 

\subsubsection{Layers}
We study if BatchEnsemble needs to be applied to all layers of a deep neural network or only in the last layers. We perform a layer-wise ablation on a 10-layer MLP, applying BatchEnsemble to the last $l$ layers. We evaluating on a synthetic regression and classification task of the Ackley function, which provides a highly non-linear benchmark that requires a deeper model to achieve good performance. Figure \ref{fig:be_layer_ablation_reg} below and Figure \ref{fig:be_layer_ablation_clf} in Appendix \ref{sec:layer_be} show that models with BatchEnsemble in all layers achieve the best predictive performance and NLL, while configurations with fewer than eight BatchEnsemble layers perform significantly worse. Additional details are provided in Appendix~\ref{sec:layer_be}.

\begin{figure}[h]
    \centering
    \includegraphics[width=0.75\linewidth]{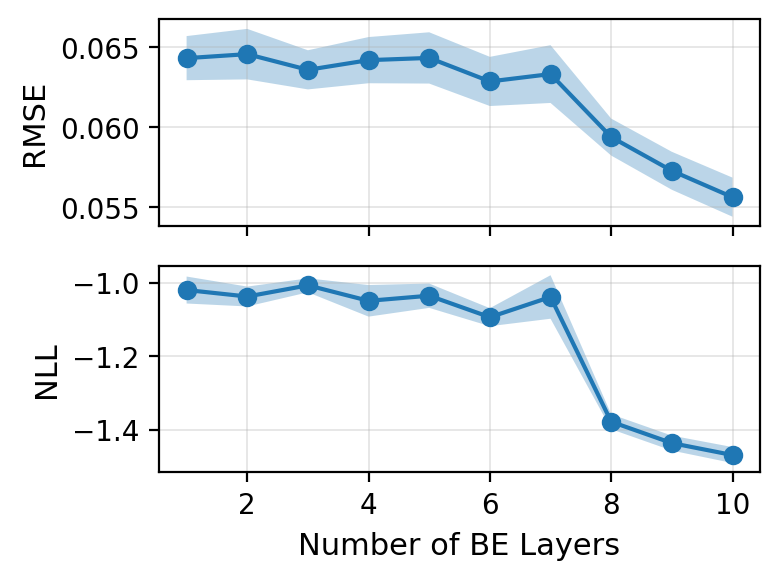}
    \vspace{-1em}\\
    \caption{Layer-wise BE ablation on the Ackley regression.}
    \label{fig:be_layer_ablation_reg}
\end{figure}

\section{Discussion and conclusion}
The overall picture is consistent across all datasets and modalities: BatchEnsemble achieves uncertainty estimates that are comparable to deep ensembles, while being significantly more efficient in terms of parameters and computation. MC dropout struggles to compete with them. 

This is in contrast to recent results for image data \cite{zamyatin2026}, who find that BatchEnsemble clearly underperforms deep ensembles in accuracy and uncertainty estimation. Our hypothesis is that parameter-sharing in CNNs makes it difficult for BatchEnsemble to learn diverse ensembles, which is not the case for tabular and time series data. This hypothesis is supported by our ablation studies that show that more BatchEnsemble layers and more BatchEnsemble adapters and GRU gates lead to better performance. However, it does not explain all the results from \citet{zamyatin2026}, who also find underperformance of a feedforward neural network on MNIST data.

An important observation is that BatchEnsemble performs well across all evaluation criteria, including calibration metrics, NLL, and selective evaluation. In particular, the selective results show that BatchEnsemble produces uncertainty estimates that meaningfully rank samples by prediction difficulty. This indicates that the uncertainty is not only well scaled, but also informative for downstream decision-making, a key requirement for practical deployment.

Under distribution shift BatchEnsemble behaves like deep ensemble. Predictive performance degrades gracefully, and uncertainty increases. 
However, uncertainty decomposition shows that BatchEnsemble and deep ensemble differ in how epistemic uncertainty is expressed across datasets. Deep ensemble exhibits the strongest epistemic increase on shifted California dataset, but BatchEnsemble shows higher epistemic uncertainty on some classification datasets. In particular, BatchEnsemble has the largest epistemic component on Breast Cancer (in-distribution and shifted) and a larger epistemic increase than deep ensembles on Phoneme. Despite these differences, the two methods produce similar predictive distributions and achieve comparable performance on all uncertainty metrics.

On time-series forecasting tasks, GRUBE performs comparable to or better than deep ensembles in predictive accuracy and uncertainty estimation. As recurrent models are sensitive to model capacity and training stability, perturbations applied at each time step may allow ensemble members to represent different temporal patterns, while shared parameters act as a regularizer and improve training stability.

\textbf{Conclusion.} 
BatchEnsemble is a scalable and general alternative to deep ensemble that provides reliable uncertainty estimates for tabular and time-series models. 

\subsection*{Acknowledgements}
We thank Anton Zamyatin and Yusuf Sale for insightful discussions.

\clearpage

\bibliography{references}
\bibliographystyle{plain}

\newpage
\appendix
\onecolumn

\section{Dataset Details}

\begin{table*}[!h]
\centering
\small 
\begin{adjustbox}{max width= 0.99\textwidth}
\begin{tabular}{l c c c p{9cm}} 
\hline
\textbf{Dataset} & Type & $\mathbf{n\times d}$ & Task &\textbf{Description} \\
\hline
\texttt{Adult} & tabular & $48,842 \times 14$ & classification & \footnotesize{Predict whether annual income of an person exceeds \$50K/yr based on census data.} \\
\texttt{Breast\_cancer} & tabular & $569 \times 30$ & classification & \footnotesize{Predict whether a person has breast cancer.} \\
\texttt{Phoneme} & tabular & $5404 \times 6$ & classification & \footnotesize{Predict whether a nasal or oral vowels.} \\
\texttt{Diabetes}  & tabular & $442 \times 10$ & regression & \footnotesize{Predict diabetes progression one year later.} \\
\texttt{California} & tabular & $20,640 \times 8$ & regression & \footnotesize{Aggregated housing data in the California district using one row per census block group.} \\

\texttt{Electric}  & time series & $397 \times 1$ & regression & \footnotesize{Monthly electricity production in the U.S from Jan. 1985 to Jan. 2018.} \\
\texttt{Temperature}  & time series & $3650 \times 1$ & regression & \footnotesize{The daily minimum temperature in Melbourne for the years 1981–1990.} \\

\hline
\end{tabular}
\end{adjustbox}
\caption{Datasets used in our experiments. Sources: UCI Machine Learning Repository \citep{adult,breast_cancer} (Adult, Breast\_cancer); Least Angle Regression \citep{Efron2004} (Diabetes); Sparse Spatial Autoregressions \citep{Pace1997} (California); FRED \citep{electricdata} (Electronic); Australian Bureau of Meteorology (Temperature). Dimensions $n\times d$ are samples$\times$features before preprocessing (one-hot encoding and missing-value handling).}

\label{tab:data_overview}
\end{table*}

\subsection{Construction of Distribution-Shifted Evaluation Sets}
\label{appendix:tails}

To create the distribution shifted data sets we create an initial base split of the data, which we fit a generalized linear model. Based on this model, we select the two most influential numerical features.  

For each selected feature, tail regions are defined as observations in the lower or upper 2.5\% quantile , computed from the training portion of the base split. A sample is considered out-of-distribution if it contains at least one tail value in any of the selected features. Assuming independence, the probability of a sample lying in a tail region is
$$
p_{\text{tail}} = 1-(1-2q)^{d_\text{selected}},
$$
with $q=0.025$ and $d_\text{selected} = 2$. The base split size is adjusted accordingly to ensure that the expected size of the final test set is approximately 20\%, $$\text{base split} = (0.2 - p_{\text{tail}}) / (1 - p_{\text{tail}}).$$

This results in training set containing only in-distribution samples and a test set that includes both in-distribution and out-of-distribution samples.

\section{Implementation Setup}
\label{appendix:setup}

For all tabular classification and regression experiments, we use a multilayer perceptron with two hidden layers of size 32. Models are trained for 500 epochs using Adam optimizer with learning rate 0.005, batch size 64, dropout rate 0.1, and no weight decay.

For time-series forecasting, we use a recurrent model with one recurrent layer followed by a two-layer MLP with hidden size 32. Training settings are kept identical to the tabular experiments.

For both deep ensemble and BatchEnsemble, we use an ensemble size of 10.

\begin{table}[h]
\centering

\begin{tabular}{lccc}
\hline
\textbf{Task / Dataset} & \textbf{BatchEnsemble} & \textbf{MC dropout} & \textbf{Deep ensemble} \\
\hline
\multicolumn{4}{l}{\textbf{Classification}} \\
\hline
\texttt{Adult} & 7,584 & 4,610 & 46,100 \\
\texttt{Breast\_cancer} & 4,308 & 2,114 & 21,140 \\
\texttt{Phoneme} & 3,258 & 1,314 & 13,140 \\
\hline
\multicolumn{4}{l}{\textbf{Regression}} \\
\hline
\texttt{California} & 3,704 & 1,410 & 14,100 \\
\texttt{Diabetes} & 3,788 & 1,474 & 14,740 \\
\hline
\multicolumn{4}{l}{\textbf{Time Series}} \\
\hline
\texttt{Electric} & 10,790 & 5,538 & 55,380 \\
\texttt{Temperature} & 10,790 & 5,538 & 55,380 \\
\hline
\end{tabular}
\caption{Parameter counts across model types, grouped by task.}
\label{tab:params_size}
\end{table}

\section{Does Orthogonal Initialization Improve Performance?}
\label{appendix:init}
Having uncorrelated members is important in ensemble methods. Effective ensembles balance two factors: strong members and low correlation among their errors. Breiman’s analysis shows that generalization improves when members are accurate and decorrelated. In tree ensembles, this is typically achieved through bagging and feature randomness \cite{Breiman2001}. 

For deep ensembles, however, bagging can reduce performance because each model trains on fewer unique examples, while random initialization and data shuffling already induce useful diversity \cite{Lakshminarayanan2017, fort2020}. For BatchEnsemble, we propose a different method of creating ensemble variation by using orthogonal initialization per member adapter $r_k$, $s_k$, and $b_k$. This de-correlates the ensemble members and reduces initial prediction correlation.

Much work has been done on orthogonal initialization in deep learning, showing how training speeds up and stabilizes \cite{hu2020}. However, not much work has been done on orthogonal weights in ensemble methods. We hypothesize that orthogonal initialization, possibly combined with orthogonal regularization, leads to less correlated ensemble members and thus improves ensemble diversity.

To test whether this improves the ensemble, we compare BatchEnsemble with orthogonal initialization to the random sign initialization from \citet{Wen2020}. We also evaluate with different strength of orthogonal regularization.

\begin{figure}[h]
    \centering
    \includegraphics[width=0.65\linewidth]{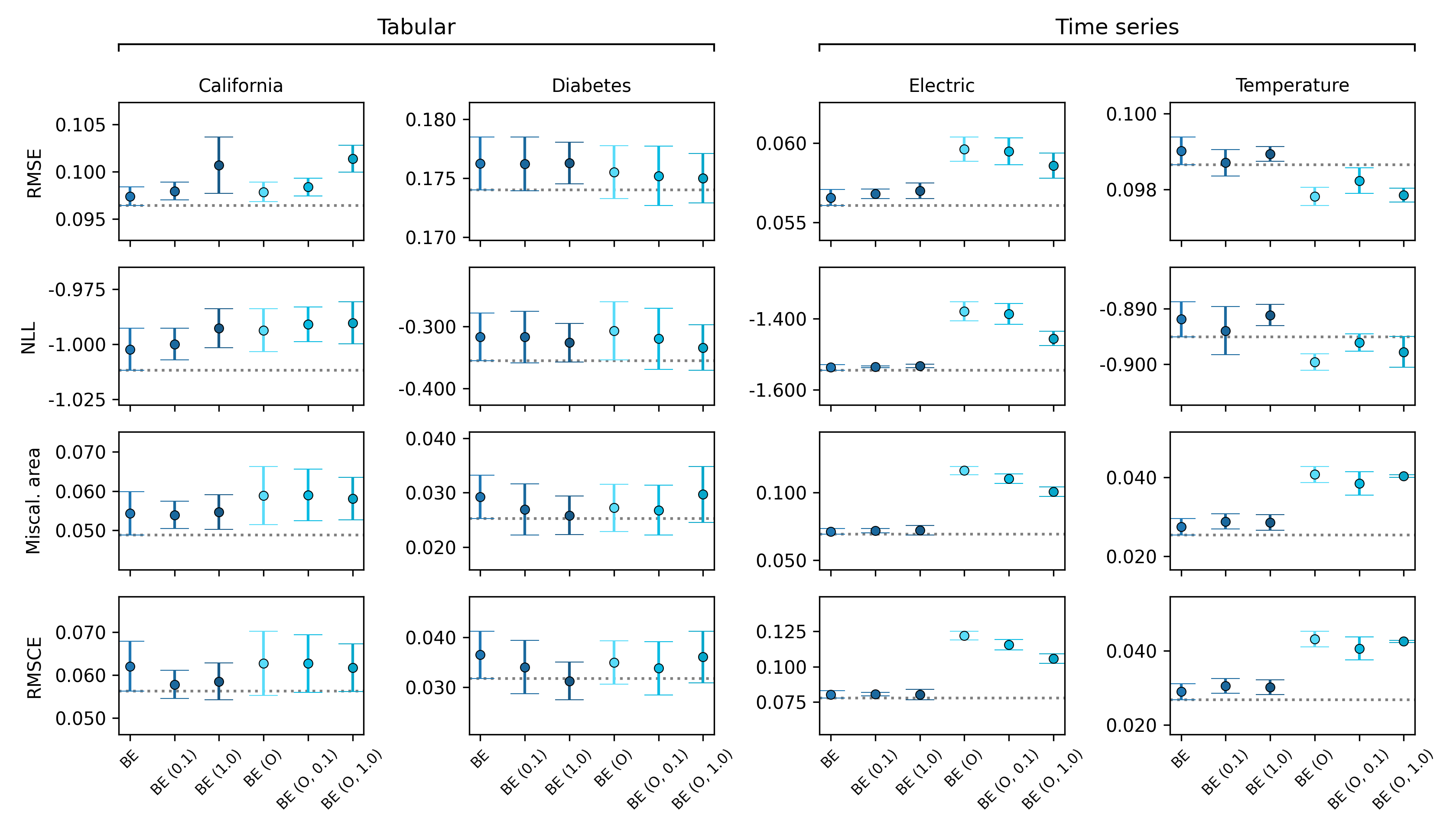}
    \vspace{-1em}\\
    \caption{Initialization and regularization ablation on regression datasets. BE uses random sign initialization. BE ($\lambda$) applies orthogonal regularization with strength $\lambda$. BE (O) uses orthogonal initialization, and BE (O,$\lambda$) combines orthogonal initialization and regularization.}
    \label{fig:be_init_ablation_reg}
\end{figure}

\begin{figure}[h]
    \centering
    \includegraphics[width=0.65\linewidth]{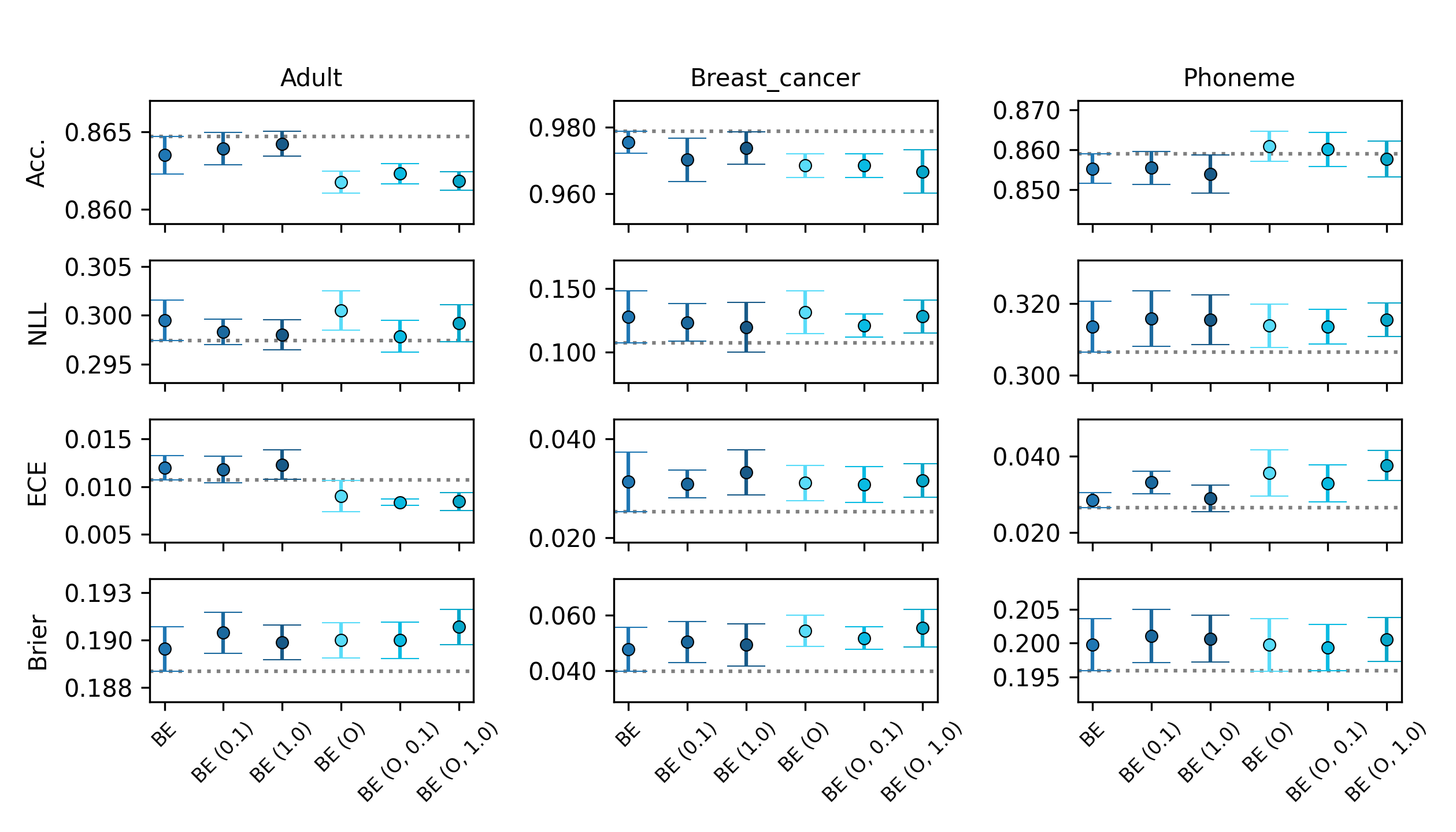}
    \vspace{-1em}\\
    \caption{Initialization and regularization ablation on classification datasets. BE uses random sign initialization. BE ($\lambda$) applies orthogonal regularization with strength $\lambda$. BE (O) uses orthogonal initialization, and BE (O,$\lambda$) combines orthogonal initialization and regularization.}
    \label{fig:be_init_ablation_class}
\end{figure}

\clearpage

\section{Do We Need All BatchEnsemble Adapters?}
\label{appendix:adapters}
The results of the adapter-wise ablation on classification datasets is shown in the main paper. Here, we show the results for the regression datasets. Again, none of the reduced models consistently match the full BatchEnsemble model.

\begin{figure}[h]
    \centering
    \includegraphics[width=0.65\linewidth]{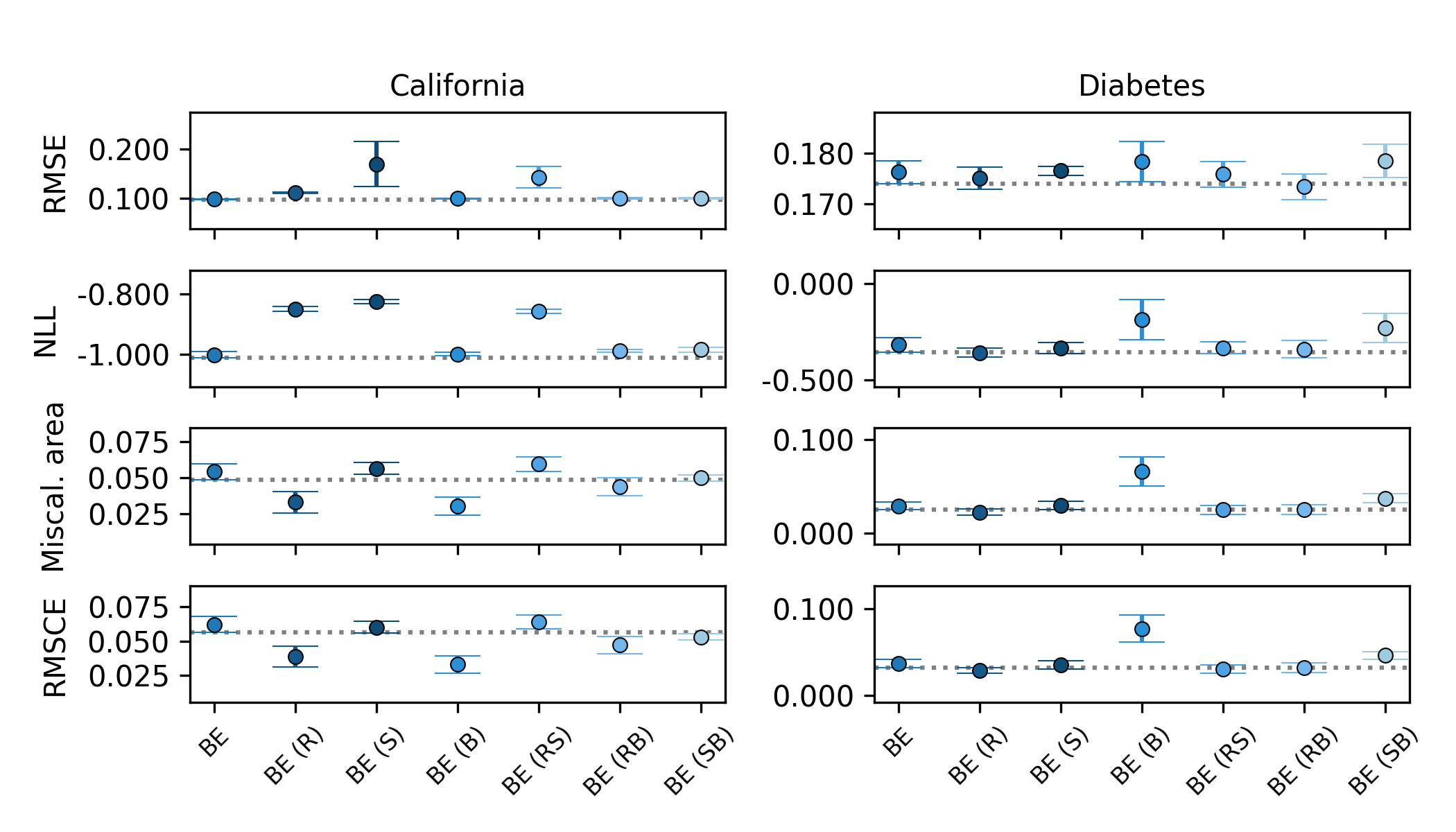}
    \vspace{-1em}\\
    \caption{Adapter-wise ablation on regression datasets.}
    \label{fig:adapter_ablation_reg}
\end{figure}

\section{Do We Need BatchEnsemble in All GRU Gates?}
\label{appendix:gru}
In GRUBE, BatchEnsemble is applied to the linear transformations of the update gate ($Z$), reset gate ($F$), and candidate state ($C$). While applying BatchEnsemble to all gates maximizes per-member flexibility, it is unclear whether ensemble diversity is required uniformly across all recurrent components, or whether uncertainty can be captured by modulating only a subset of gates.
To study this, we perform a gate-wise ablation in which BatchEnsemble layers are selectively applied to different combinations of gates. We evaluate the configurations ${\text{C}, \text{Z}, \text{F}, \text{CZ}, \text{CF}, \text{ZF}, \text{CZF}}$, where $\text{CZF}$ corresponds to BatchEnsemble applied to all gates. All variants share the same recurrent architecture and training setup, differing only in which gates use BatchEnsemble parameterization. This experiment isolates the contribution of each gate to predictive performance and uncertainty estimation, and assesses whether BatchEnsemble is necessary throughout the GRU cell or only in specific components.

\begin{figure}[h]
    \centering
    \includegraphics[width=0.65\linewidth]{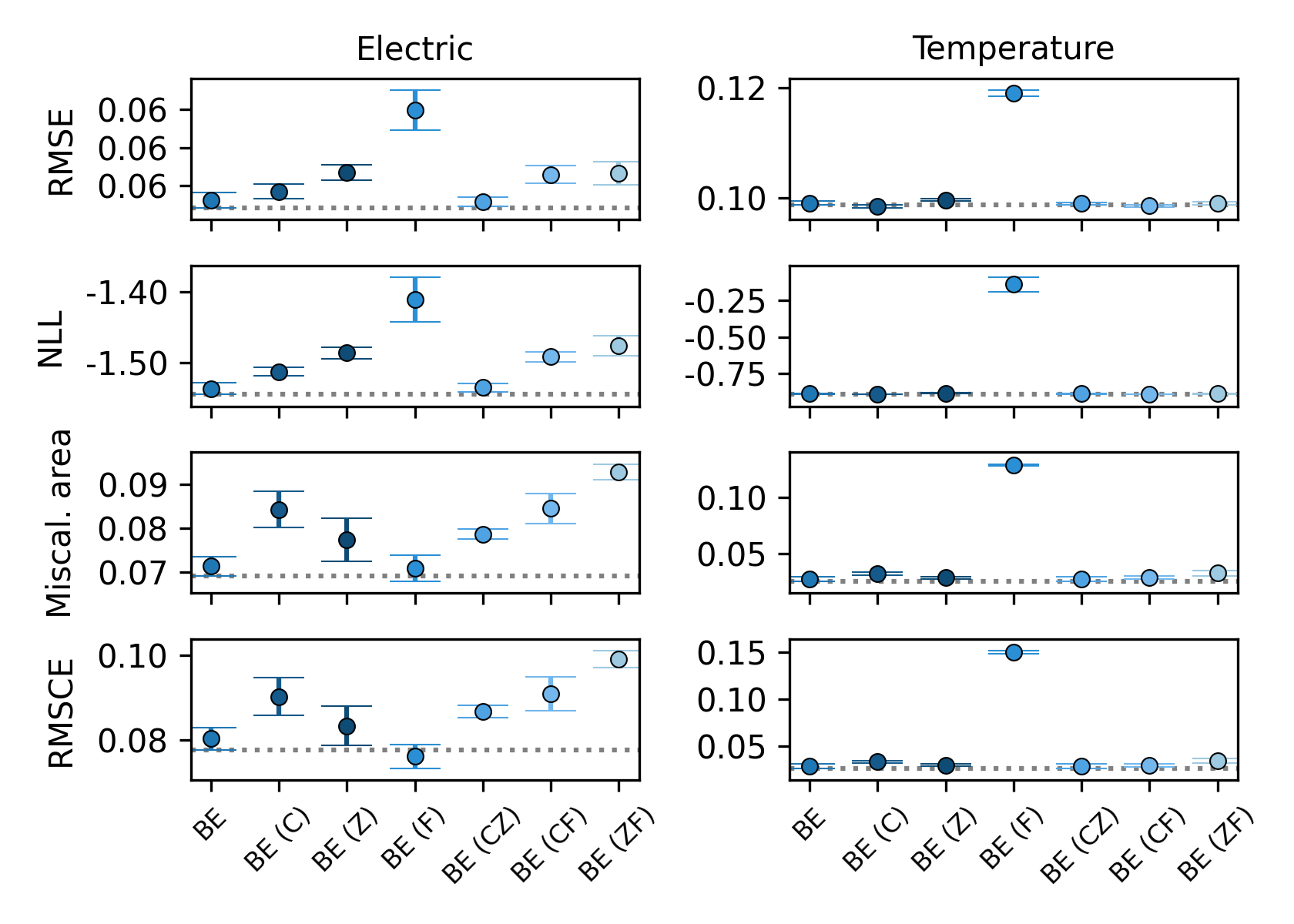}
    \vspace{-1em}\\
    \caption{Adapter-wise ablation in GRU gates.}
    \label{fig:be_gru_ablation}
\end{figure}

\section{Do We Need BatchEnsemble in All Layers?}
\label{sec:layer_be}

BatchEnsemble is commonly applied to all layers, but it is unclear whether it is required to obtain strong predictive performance and uncertainty estimates. It may be sufficient to apply BatchEnsemble only in the later layers.

To investigate this, we conduct a layer-wise ablation on a 10-layer MLP, where BatchEnsemble is applied to only the last $k \in \{1,\dots,10\}$ layers while the remaining layers are standard. All other architectural and training settings are kept fixed. This experiment evaluates whether BatchEnsemble is needed across all layers or only in the later part of the network.
These experiments were performed on regression and classification versions of the Ackley-function

\begin{equation*}
\begin{aligned}
f(\mathbf{x}) ={}&
- a \exp\!\left(-b \sqrt{\tfrac{1}{d}\sum_{i=1}^d x_i^2}\right) 
- \exp\!\left(\tfrac{1}{d}\sum_{i=1}^d \cos(c x_i)\right)
+ a,
\end{aligned}
\end{equation*}
where \(a = 20\), \(b = 0.2\), \(c = 2\pi\), and \(d=10\) denotes the input dimension. To model aleatoric uncertainty, we add heteroscedastic noise $\sigma(\mathbf{x})$ that increases with the normalized input radius \(\|\mathbf{x}\|\).

For regression, the targets are generated as
\[
y = f(\mathbf{x}) + \sigma(\mathbf{x})\,\epsilon, \quad \epsilon \sim \mathcal{N}(0,1),
\]
yielding a standard heteroscedastic regression problem. 

We construct the binary classification version of the Ackley function by We first computing a deterministic score
 $$s(x) = f(x)- \mathrm{median}(f(x)),$$
 where the median is taken over the sampled dataset. We then add heteroscedastic noise to the score and threshold at zero
 $$\tilde{s}(\mathbf{x}) = s(\mathbf{x}) + \sigma(\mathbf{x})\,\epsilon, 
\qquad
y = \mathbb{I}\{\tilde{s}(\mathbf{x}) > 0\}.$$
Points above the 0 are assigned to one class, and points below to the other. The heteroscedastic noise results in labels near the center of the domain are relatively stable, while labels farther from the origin become increasingly uncertain, giving a soft decision boundary with radially increasing ambiguity.
 
\begin{figure}[h]
    \centering
    \includegraphics[width=0.49\linewidth]{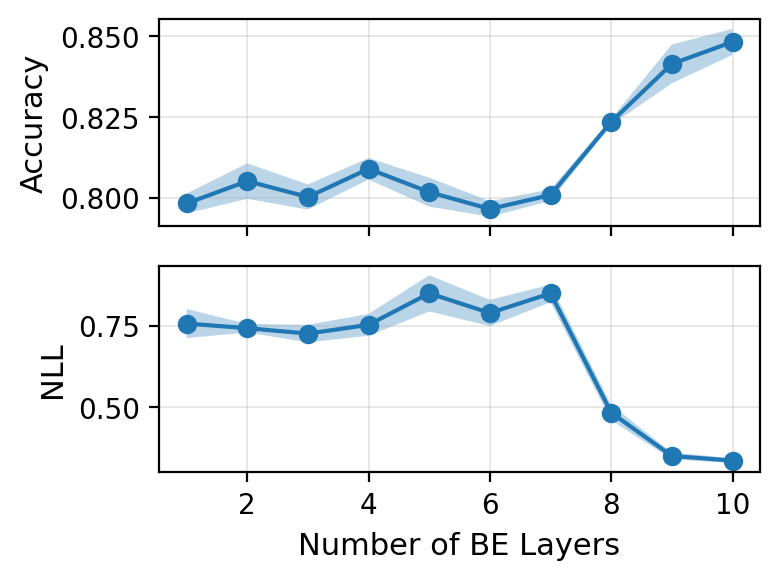}
    \vspace{-1em}\\
    \caption{Layer-wise BE ablation on Ackley classification.}
    \label{fig:be_layer_ablation_clf}
\end{figure}

\clearpage

\section{Uncertainty Decomposition}
Here we show the full uncertainty decomposition results for tabular classification. \cref{tab:unc_id_shift_delta_reg} shows the same for the regression datasets. 

\begin{table}[h]
\centering
\small

\textbf{\scriptsize Dataset: \texttt{Adult}}\\
\begin{adjustbox}{max width=0.65\textwidth}
\begin{tabular}{lccc|ccc|ccc}
\toprule
 & \multicolumn{3}{c|}{ID} & \multicolumn{3}{c|}{Shift} & \multicolumn{3}{c}{$\Delta$ (Shift $-$ ID)} \\
\cmidrule(lr){2-4} \cmidrule(lr){5-7} \cmidrule(lr){8-10}
Model & total & aleatoric & epistemic & total & aleatoric & epistemic & total & aleatoric & epistemic \\
\midrule
BatchEnsemble & 0.293 & 0.286 & 0.007 & 0.277 & 0.268 & 0.009 & -0.016 & -0.018 & +0.002 \\
MC dropout & 0.280 & 0.258 & 0.022 & 0.245 & 0.223 & 0.022 & -0.035 & -0.035 & +0.000 \\
Deep ensemble & 0.291 & 0.264 & 0.027 & 0.253 & 0.222 & 0.031 & -0.038 & -0.042 & +0.004 \\
Single & 0.266 & 0.266 & 0.000 & 0.231 & 0.231 & 0.000 & -0.035 & -0.035 & +0.000 \\
\bottomrule
\end{tabular}
\end{adjustbox}
\vspace{0.01em}\\
\textbf{\scriptsize Dataset: \texttt{Breast\_cancer}}\\
\begin{adjustbox}{max width=0.65\textwidth}
\begin{tabular}{lccc|ccc|ccc}
\toprule
 & \multicolumn{3}{c|}{ID} & \multicolumn{3}{c|}{Shift} & \multicolumn{3}{c}{$\Delta$ (Shift $-$ ID)} \\
\cmidrule(lr){2-4} \cmidrule(lr){5-7} \cmidrule(lr){8-10}
Model & total & aleatoric & epistemic & total & aleatoric & epistemic & total & aleatoric & epistemic \\
\midrule
BatchEnsemble & 0.043 & 0.027 & 0.016 & 0.034 & 0.019 & 0.015 & -0.009 & -0.008 & -0.001 \\
MC dropout & 0.032 & 0.028 & 0.004 & 0.024 & 0.017 & 0.007 & -0.008 & -0.011 & +0.003 \\
Deep ensemble & 0.033 & 0.026 & 0.007 & 0.023 & 0.014 & 0.009 & -0.010 & -0.012 & +0.002 \\
Single & 0.027 & 0.027 & 0.000 & 0.018 & 0.018 & 0.000 & -0.009 & -0.009 & +0.000 \\
\bottomrule
\end{tabular}
\end{adjustbox}
\vspace{0.01em}\\
\textbf{\scriptsize Dataset: \texttt{Phoneme}}\\
\begin{adjustbox}{max width=0.65\textwidth}
\begin{tabular}{lccc|ccc|ccc}
\toprule
 & \multicolumn{3}{c|}{ID} & \multicolumn{3}{c|}{Shift} & \multicolumn{3}{c}{$\Delta$ (Shift $-$ ID)} \\
\cmidrule(lr){2-4} \cmidrule(lr){5-7} \cmidrule(lr){8-10}
Model & total & aleatoric & epistemic & total & aleatoric & epistemic & total & aleatoric & epistemic \\
\midrule
BatchEnsemble & 0.337 & 0.328 & 0.009 & 0.378 & 0.352 & 0.026 & +0.041 & +0.024 & +0.017 \\
MC dropout & 0.334 & 0.312 & 0.022 & 0.362 & 0.328 & 0.034 & +0.028 & +0.016 & +0.012 \\
Deep ensemble & 0.323 & 0.316 & 0.007 & 0.351 & 0.334 & 0.017 & +0.028 & +0.018 & +0.010 \\
Single & 0.310 & 0.310 & 0.000 & 0.330 & 0.330 & 0.000 & +0.020 & +0.020 & +0.000 \\
\bottomrule
\end{tabular}
\end{adjustbox}
\vspace{-1em}
\caption{Uncertainty decomposition for tabular classification under in-distribution (ID) and shifted data. $\Delta$ denotes Shift $-$ ID.}
\label{tab:unc_id_shift_delta_clf}
\end{table}

\section{Selective Prediction Evaluation with Distribution Shift} \label{appendix:selective}
Here we also compare selective evaluation results under distribution shift for the tabular datasets. Results are comparable to in-distribution selective evaluation.  

\begin{figure}[h]
    \centering
    \includegraphics[width=0.65\linewidth]{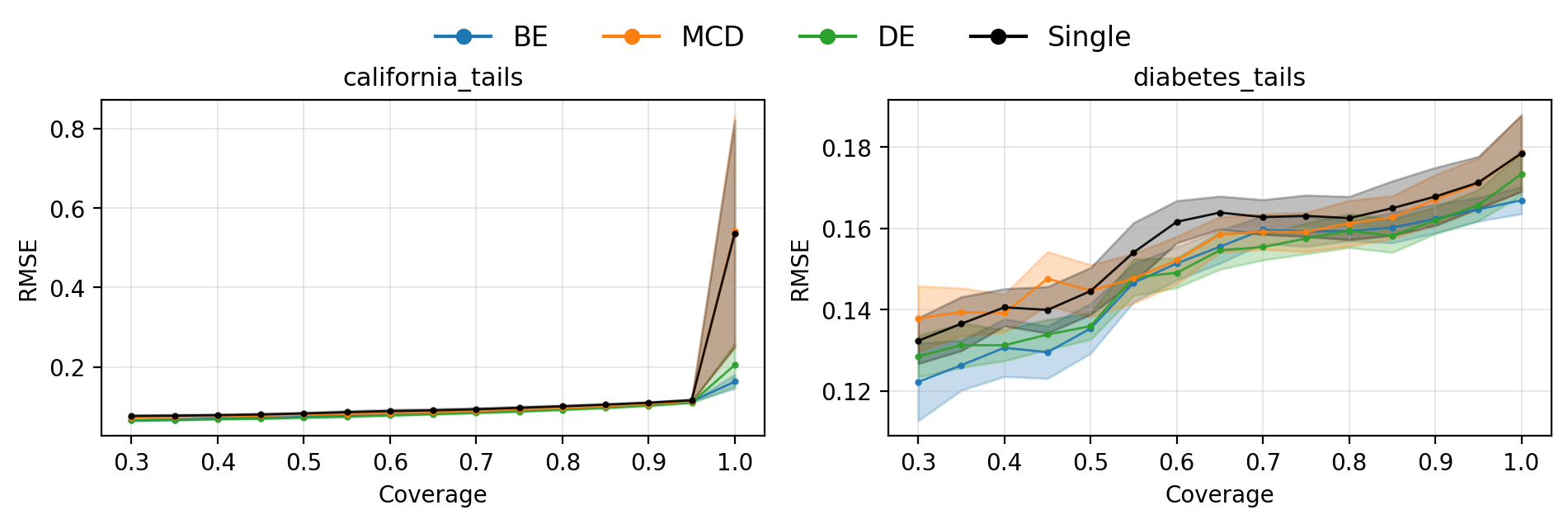}
    \vspace{-1em}\\
    \caption{Selective evaluation for tabular regression tail datasets.}
    \label{fig:selective_reg_tails_all}
\end{figure}

\begin{figure}[h]
    \centering
    \includegraphics[width=0.85\linewidth]{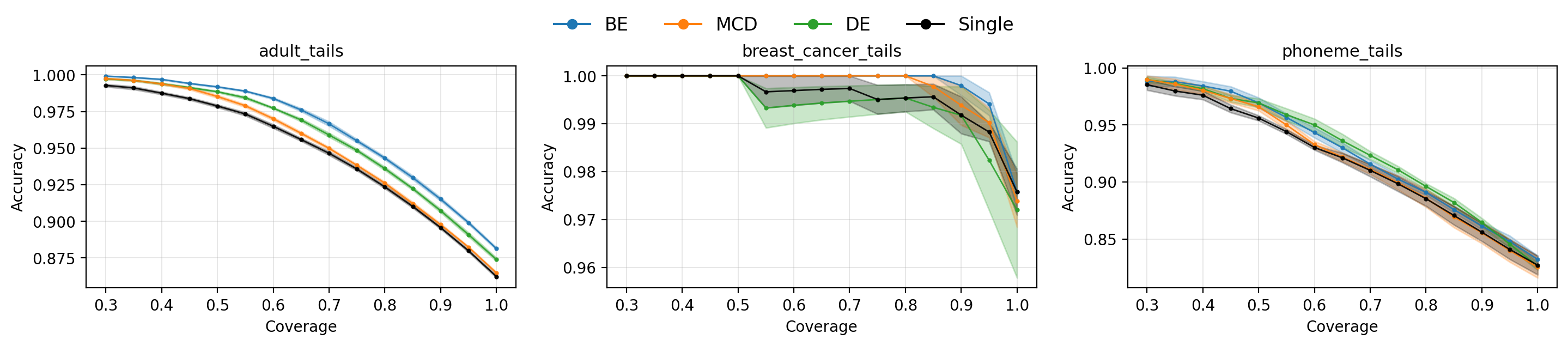}
    \vspace{-1em}\\
    \caption{Selective evaluation for tabular classification tail datasets.}
    \label{fig:selective_clf_tails_all}
\end{figure}

\clearpage
\section{Nominal vs. Empirical Coverage Results}
\label{appendix:nom_vs_emp}
Here we show that all models are reasonably well-calibrated on the regression data. All models consistently overestimate uncertainty on the California dataset and underestimate uncertainty on the diabetes dataset. 
\begin{figure}[h]
    \centering
    \includegraphics[width=0.49\linewidth]{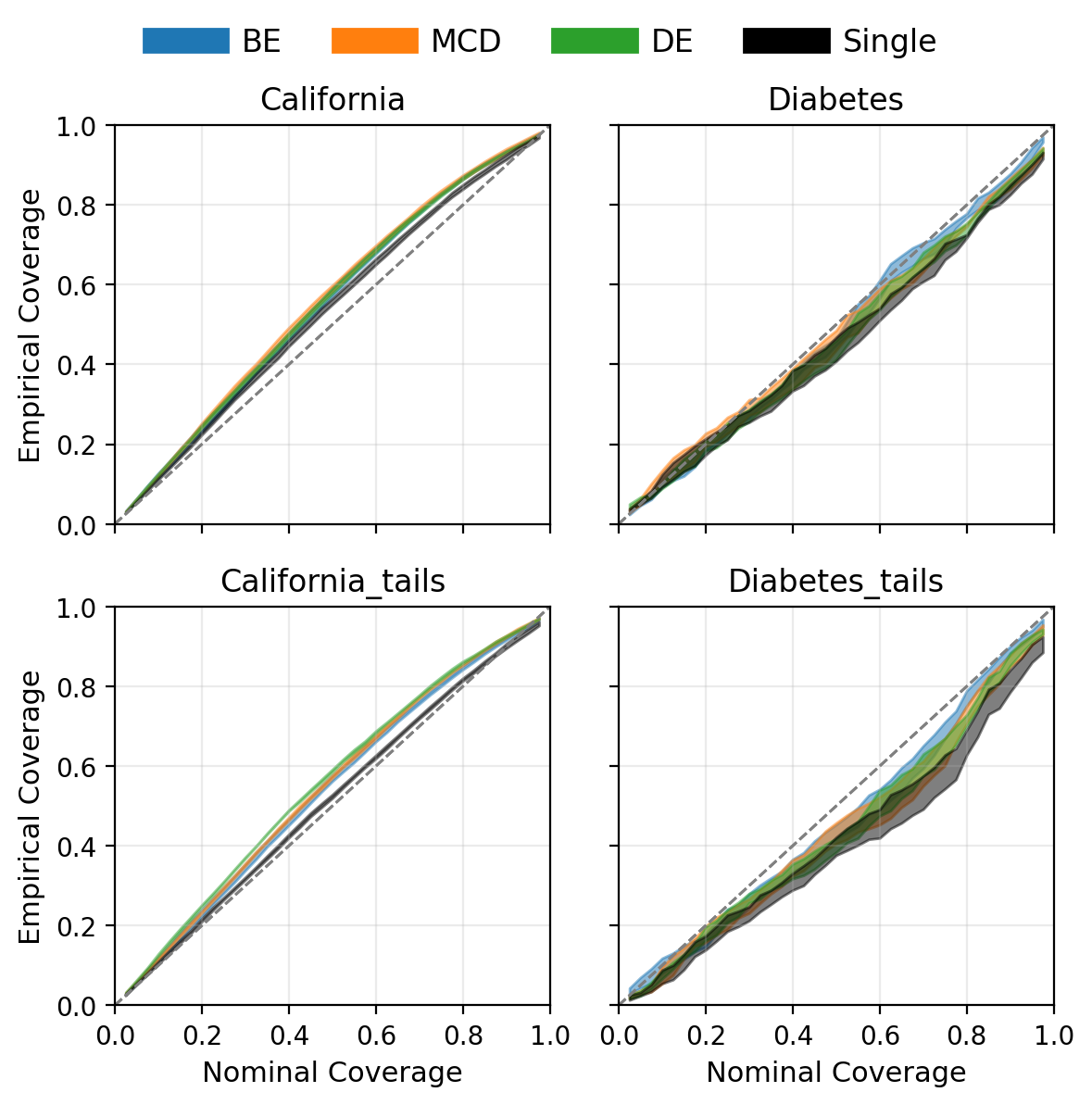}
    \vspace{-1em}\\
    \caption{Nominal versus empirical coverage for the tabular regression datasets. The dashed line denotes ideal calibration.}
    \label{fig:nom_vs_emp_cov_reg}
\end{figure}


\end{document}